# From Basic Affordances to Symbolic Thought: A Computational Phylogenesis of Biological Intelligence

**John E. Hummel**[1,3,4] **and Rachel F. Heaton**[2,3,4,5]

[1]Department of Psychology
[2]School of Art and Design
[3]Department of Philosophy
[4]Informatics
[5]Siebel Center for Design
University of Illinois Urbana-Champaign

## Author Note

John E. Hummel https://orcid.org/0000-0002-1585-9155
Rachel F. Heaton https://orcid.org/0000-0002-7121-7691

The code and other materials for these simulations have been made publicly available in the Open Science Framework repository and can be accessed at https://osf.io/p5zjw. The images produced for illustration purposes in the General Discussion are available at the same link. The simulations presented here were not preregistered. These data were shared in an online presentation to the Analogical Minds Seminar and the 2019 Minds and Machines Research Meeting at the University of Bristol. This research was supported by AFOSR Grant AF-FA9550-12-1-003. We are grateful to Alex Doumas and Jonathan Livengood for helpful discussions of the ideas presented here. We have no conflict of interest to disclose. Correspondence concerning this article should be addressed to John E. Hummel, Department of Psychology, University of Illinois, 603 E. Daniel St., Champaign, IL, 61820. Email: jehummel@illinois.edu

## Abstract

What is it about human brains that allows us to reason symbolically whereas most other animals cannot? There is evidence that dynamic binding, the ability to combine neurons into groups on the fly, is necessary for symbolic thought, but there is also evidence that it is not sufficient. We propose that two kinds of *hierarchical integration*—integration of multiple role-bindings into multi-place predicates, and integration of multiple correspondences into structure mappings—are minimal requirements, on top of basic dynamic binding, to realize symbolic thought. We tested this hypothesis in a systematic collection of 17 simulations that explored the ability of cognitive architectures with and without the capacity for multi-place predicates and structure mapping to perform various kinds of tasks. The simulations were as generic as possible, in that no task could be performed based on any diagnostic features, depending instead on the capacity for multi-place predicates and structure mapping. The results are consistent with the hypothesis that, along with dynamic binding, multi-place predicates and structure mapping are minimal requirements for basic symbolic thought. These results inform our understanding of how human brains give rise to symbolic thought and speak to the differences between biological intelligence, which tends to generalize broadly from very few training examples, and modern approaches to machine learning, which typically require millions or billions of training examples. The results we report also have important implications for bio-inspired artificial intelligence.

Human thinking appears to differ qualitatively from the cognitive abilities of the other great apes (Penn et al., 2008). Unlike our closest primate cousins (bonobos, chimpanzees, orangutans, and gorillas) the human species has developed language, mathematics, science, engineering, economics, a knowledge of our own history, and everything they make possible. As a result, we are by some criteria the dominant species on the planet. The seeming differences between human cognition and the cognitive abilities of other animals have been attributed to the human cognitive architecture implementing a *symbol system*: A computing architecture capable of composing a finite number of atomic elements (neurons, words, variables, values, relations, etc.) into a much larger set of expressions, and acting on those expressions (e.g., Boole, 1847, 1854; Chomsky, 1972; Deacon 1997; Fodor & Pylyshyn, 1988; Peirce, 1897, 1903; Penn, et al., 2008; Quilty-Dunn et al., 2023). If the symbol system in question is capable of recursion, with expressions taking other expressions as arguments, then the number of possible expressions becomes unbounded and it becomes possible to express ideas that would not have been expressible otherwise.

This *symbol system hypothesis* is not uncontroversial. Some in the connectionist (e.g., Rumelhart, et al., 1986; Doerig et al., 2023), behaviorist (e.g., Skinner, 1957), and dynamical systems (e.g., Abraham et al., 1990) communities have historically questioned the need for symbol systems. Many in the comparative cognition literature also argue that the difference between human cognition and that of other great apes is a matter of degree rather than kind (e.g., Tomasello & Call, 1997).

It is not our purpose to defend the empirical merits of the symbol system hypothesis or the claim that human cognition differs qualitatively from the cognitive abilities of other animals. For that, we refer the reader to Deacon (1997), Penn et al. (2008), and the references therein. Nonetheless, there are interesting findings in the animal cognition literature that are worth highlighting because they raise the question of whether or to what degree nonhuman animals might use or at least approximate symbolic processes in reasoning. For example, wild bumblebees can bind colors and shapes into representations of objects and respond to relations between objects (Loukola et al., 2017; Martin-Ordas, 2023) and honeybees can perform match-to-sample tasks (Martin-Ordas, 2022; Moreno et al., 2012; Ng et al., 2020). Rats can infer causal relations that go beyond mere associations (Blaisdell, et al., 2006). Chimpanzees are well known to make and use tools (Boesch & Boesch, 1990; however, as demonstrated by Povinelli, 2003, it is unlikely that they understand why the tools work). Chimpanzees can also solve match-to-sample problems, and symbol-trained chimps can even solve relational match-to-sample problems (Thompson et al., 1997). Corvids (i.e., crows, ravens, jays, and magpies) can use and make tools (Bird & Emery, 2009; Claton & Emery, 2005; Giri & Garcia-Pelegrin, 2025) and engage in multi-step problem solving (Bird & Emery, 2009). Orcas are known to hunt cooperatively (Pitman & Durbin, 2012). As impressive as these capacities are, it remains unclear which of them require fully symbolic representations and processes.

Our focus in this paper is the question of what it would mean for a neural computing architecture such as a brain to implement a symbol system and the related question of what makes it difficult for them to do so. In the roughly two-billion-year history of multicellular life on Earth, symbolic cognition seems to have appeared between about a hundred thousand and a million years ago. If the history of multicellular life on Earth were compressed into a single day, then symbolic thought has only existed for at most the last 43 seconds, and perhaps as little as the last four seconds. Given the adaptive value of symbolic thought, an obvious question is *What took so long?* What makes achieving a symbol system in a neural computing architecture so difficult?

The simulations we will present suggest that the answer to this question might be that the leap from pre-symbolic to the most basic form of symbolic cognition requires two distinct mutations to occur in the same organism: A capacity to hierarchically combine multiple role bindings into multi-place predicates, and a capacity to compute mappings between systems of predicates. Prior work on analogical reasoning suggests that a third capacity—namely, the ability to infer entire relational structures from one system into another as a consequence of the mappings between them, a process known as *Copy with Substitution and Generation* (CWSG; Gentner, 1983; Gick & Holyoak, 1980, 1983; Holyoak & Thagard, 1989)—is necessary for the most complex forms of symbolic reasoning. Students of analogical reasoning have long identified analogical inference, the process of making an inference about a target problem based on its mapping to an analogous source, with the process of CWSG, in which elements in the source are inferred (*copied*) into the target, *substituting* them with existing elements in the target when possible and *generating* those elements in the target (from the source) when necessary (see Hummel & Holyoak, 2003). To our knowledge, all models of analogical inference incorporate this process in their operation (e.g., Falkenhainer et al., 1989; Holyoak & Thagard, 1989; Hummel & Holyoak, 2003).

However, CWSG is a comparatively small algorithmic advance over the capacity for mapping in that it requires only the ability to detect when all elements of the target are inhibited by the source to target mapping (see Hummel & Holyoak, 2003). Armed with CWSG, Hummel and Holyoak demonstrated that schema induction is likewise a simple matter of being able to detect when elements of the source are shared by the target (i.e., to represent the intersection of the two; see also Doumas et al., 2008, 2022). As such, we view mapping and the capacity to form multi-place predicates, more than CWSG and intersection discovery, to be the major hurdles to achieving symbolic thought in a neural computing architecture. Given this, a second focus of this paper is to understand what each of these abilities in isolation can do for an organism.

## Symbol Systems, Relational Reasoning, and Neural Computation

Penn et al. (2008) summarize evidence that human cognition differs from the cognitive abilities of other great apes in our capacity for *role-based relational reasoning* (RRR), a manifestation of our capacity for symbolic thought. RRR is the ability to make inferences and generalizations based, not on the features of objects, but on the relational roles to which they are bound. For example, if one knows that $x$ is larger than $y$ and that $y$ is larger than $z$, then one can infer that $x$ is larger than $z$, even if one has no idea what $x$, $y$, and $z$ are. This inference is based strictly on the bindings of $x$, $y$, and $z$ to the roles of the *larger-than* () relation, rendering it independent of the objects' other properties. RRR thus leads to universal inductive generalization, including to objects and situations that are completely unfamiliar or even unspecified (as in the case of $x$, $y$, and $z$). This capacity for universal generalization is the basis of logic, mathematics, and everything that depends on them, including science, engineering, trade, language, our understanding of our history, and all of humankind's most notable achievements. In turn, RRR—and the symbol system on which it relies—depends at its core on the ability to bind values to variables or equivalently arguments to relational roles.

The first obstacle to achieving a symbol system in a neural computing architecture is that the format of the bindings matters (Hummel, 2011; Hummel & Biederman, 1992; Hummel & Holyoak, 1997, 2003; Shastri & Ajjenagadde, 1993). To support universal inductive generalization, the bindings must be *dynamic*, meaning that they can be created and destroyed on the fly and that the representation of the relational roles (variables) and their arguments (values) are unaffected by the binding process. The same symbol (e.g., collection of neurons) must represent the roles of *larger-than* (*x, y*) in the expression *larger-than* (lion, monkey) as in the expression *larger-than* ($2,000, $1,000); and the same symbols (neurons) must represent *lion* and *monkey* in the expression *larger-than* (lion, monkey) as in the expression *preys-on* (lion, monkey). Otherwise, the system would fail to capture what the expressions have in common and how they differ. It would not be a symbol system. In the words of Fodor and Pylyshyn (1988), it would lack *systematicity*.

One approach to achieving dynamic binding in artificial neural networks, first proposed by von der Malsburg & Buhmann (1982) and later adopted by many others (e.g., Doumas et al., 2008, 2022; Heaton & Hummel, 2019; Hummel & Biederman, 1992; Hummel & Holyoak, 1997, 2003; Shastri & Ajjanagadde, 1993; Strong & Whitehead, 1989), is to use synchrony of neural firing. The basic idea is that neurons that are bound together fire in synchrony (or close asynchrony; Doumas et al., 2008, 2022) with one another and out of synchrony with neurons representing other bindings. For example, to represent *larger-than* (lion, monkey), the neurons representing the *larger* role would fire in synchrony (or close asynchrony) with the neurons representing *lion*, and out of synchrony (or in greater asynchrony) with the neurons representing *smaller* and *monkey*.[1] To represent *larger-than* (monkey, lion), the same neurons would be active but their synchrony relations would be reversed.

**Beyond Dynamic Binding**

The second obstacle to achieving a symbol system in a neural computing architecture is that although dynamic binding is necessary for a symbol system, it is not sufficient. Synchrony of neural firing of the kind required for dynamic binding has been observed in the brains of monkeys (e.g., Ardestani et al., 2016; Fuster, 2009; Tallon-Baudry et al., 2004), cats (Gray & Singer, 1998), and even insects (Batheiller et al., 2010; Galizia & Sachse, 2010; Stopfer, 2007; Tanaka et al., 2009; Turner et al., 2007), none of which appears capable of fully symbolic thought. More importantly, many animals appear capable of universal inductive generalization in specific domains, such as discovering causal relations (Blaisdell et al., 2006) and spatial navigation. Housecats, for example, are expert at exploring spaces they have never seen before. Without any prior experience with a space, a cat knows immediately which surfaces afford walking or sitting, which spaces afford hiding, which are likely to hide small creatures of interest, and which objects on the shelf afford knocking to the floor. This observation suggests that although symbolic thought might be necessary to master mathematics, science, and engineering, it is not necessary for successful interaction with novel environments, including universal inductive generalization of basic perceptual affordances (Gibson, 1977) or causal relations (Blaisdell et al., 2006). In this way, most animals evidence a capacity for generalization that exceeds the capacity of any simple associative learning mechanism.

---

[1] The essential role of dynamic binding for achieving a symbol system in a neural computing architecture does not imply that there is no role for *static* binding (e.g., by conjunctive coding, in which a neuron codes explicitly for a conjunction of, say, a role and its argument). As elaborated later, static binding is necessary for storing bindings in long-term memory. But the resulting static bindings, by themselves, are not sufficient to render a neural representation a symbol system (see Hummel, 2011; Hummel & Holyoak, 1997, 2003).

**The Hypothesis**

If dynamic binding is necessary but not sufficient for symbolic thought, then what is the difference between a symbolic species and a non-symbolic one? What is the neurocomputational difference between being able to interact with a novel environment (e.g., knowing how to navigate through it) and being able to control it (e.g., by building a home, or extracting electric power from a river)?

Our hypothesis is that, in addition to dynamic binding, the minimal requirement for the most basic kind of symbolic processing is *hierarchical integration*: The capacity to compose individual role bindings into multi-place relational structures and the capacity to map such structures onto one another. It is one thing to dynamically bind individual values (e.g., *this surface*, or *this object*) to individual variables (e.g., *surfaces that afford walking on* or *objects small enough to push around*), but it is more difficult to (a) integrate multiple such variable-value or role-argument bindings into complete, recursive, relational structures, and (b) map systems of such structures onto one another in the service of using one to reason about the other.

**Single- vs. Multi-place Predicates**

It is straightforward to *predicate*—that is, to represent as an explicit predicate—that some property, $p$, applies to an object. As illustrated under Simulations, all that is required is the ability to (a) represent $p$ independently of the object's other properties and (b) dynamically bind the representation of $p$ to a representation of the object's other properties. For example, to represent that some surface, $s$, affords walking on, a cat need only represent the predicate *can-walk-on* () and the argument $s$ and dynamically bind the predicate to its argument. The resulting representation, *can-walk-on* ($s$), informs the cat that it is possible to walk on $s$.

The same principle applies to the affordances the dorsal visual system represents in the service of our interactions with objects in the world (see Goodale & Milner, 1992). To grasp an object such as a glass of water, we need to know the distance between the glass and our hand, *distance* (glass, hand), the direction in 3-space between the glass and our hand, *direction* (glass, hand), and the shape and size of the glass relative to our hand, *shape* (glass, hand). This kind of information is coded in the *posterior parietal* region of our visual cortex (area PP), which resides very close to motor cortex (Goodale & Milner, 1992).

Considerations such as these have led some to conclude that PP represents the spatial relations between objects in our environment. And in a sense, that's what they do. But in another sense, it's not what they do. In all these egocentric relations—*distance* (glass, hand), *angle* (glass, hand), *shape* (glass, hand)—one of the arguments, namely "hand", is fixed. It may not be a glass we wish to grasp (perhaps it's a fork or a computer mouse), but in every case, it is our hand that will do the grasping. As a result, all these "relations" can be reduced to single-place predicates—*hand-distance* (glass), *hand-angle* (glass), *hand-shape* (glass)—simply by dropping the argument that remains constant. In this sense, the "relations" represented in PP may be degenerate (i.e., relations in a formal sense, but not represented as such) because they can be reduced to single-place predicates without any meaningful loss of information. The only meaningful cost is that separate sets of predicates (neurons) are required to represent objects relative to each body part (e.g., with one set for the hand, another for the eyes, the head, etc.). But given that the number of relevant body parts is comparatively small, such a coding scheme would be both feasible and serviceable for an organism without the capacity to represent multi-place predicates. And it might even be an efficient solution for an organism with that capacity, since it would require only half as many role-bindings as an explicitly relational representation.

In contrast to egocentric relations such as those represented in PP, allocentric spatial relations between objects in the world, for example as represented in the lateral occipital complex (LOC; Kim & Biederman, 2011), cannot be reduced to single-place predicates because none of the arguments are fixed. For example, to represent whether one object, $x$, will fit inside another object, $y$, one must be able to represent the multi-place relation *can-fit-inside* $(x, y)$, where $x$ and $y$ are both free to vary.

As such, representing a single-place predicate such as *can-walk-on* $(x)$ or *hand-distance* $(x)$ is qualitatively different than representing a multi-place predicate such as *can-fit-inside* $(x, y)$. The former only requires a cognitive architecture to dynamically bind a representation of the variable/role (e.g., *can-walk-on*) to a representation of its value (e.g., *this-surface*)—a problem we have neurophysiological and behavioral evidence to suggest many animals can solve. The latter requires the cognitive architecture to represent multiple role bindings at the same time and integrate them into a single hierarchical structure. Representing multi-place relations such as *larger-than* $(x, y)$ depends on the capacity to hierarchically compose multiple role-bindings such as *larger* $(x)$ and *smaller* $(y)$ into multi-place predicates. Without a capacity to hierarchically combine multiple role bindings, it is therefore impossible to represent the open-ended class of relations that characterizes productive symbolic thought and RRR.

**Multi-place predicates, recursion, and unbounded compositionality**: The capacity to hierarchically compose multiple role-bindings into multi-place predicates also confers another advantage associated with symbol systems, namely unbounded compositionality. With dynamic binding alone—that is, with only single-place predicates—it is possible, in the limit, to compose $n$ representational elements (neurons, symbols, etc.) into $2^n$ bindings. That is, the number of possible bindings (i.e., subsets) of $n$ elements is the powerset of $n$—a number that, for large $n$ is enormous but finite.

By contrast, with multi-place predicates it is possible to have *recursion*, in which complete structures can serve as arguments of other structures. For example, such predicates can represent not only that an atomic nucleus is larger than an electron, *larger-than* (nucleus, electron), but also that Mary knows this fact, *knows* (Mary, *larger-than* (nucleus, electron)), and that Mary's knowledge allows her to infer other properties of atoms, *allows* (*knows* (Mary, *larger-than* (nucleus, electron)), *infer* (Mary, $f$ (…))).[2] With recursion over multi-place predicates, the expressive power of a representational system with a vocabulary of $n$ elements goes from the large-but-finite powerset, $2^n$ expressions, to an unbounded number of expressions.

By conferring the capacity to represent an open-ended vocabulary of relations and form an infinite number of expressions from a finite vocabulary of elements, the ability to compose multiple role-bindings into hierarchical, multi-place predicates changes a representational system qualitatively. It renders the system symbolic. The capacity for recursion also qualitatively changes the kinds of ideas a representational system can express. Without recursion, it is impossible to express ideas such as "I think $x$", "I disagree with $x$", "It used to be the case that $x$", etc.

---

[2] It is of course possible to compose embedded structures with single-place predicates, for example *large* (*red* (*sittable* (surface))), but such embeddings are formally equivalent to "flat" (nonhierarchical) predicates, in this case *large-red-sittable* (surface), in the sense that the binding of *large*, *red*, and *sittable* to "surface" is simply one subset of the $n$ representational elements that include *large*, *red*, *sittable*, and "surface": This bound subset of the $n$ elements is already part of the powerset of $n$.

**Structure Mapping**

Although integrating role bindings into multi-place predicates turns a representational system into a symbol system, the capacity to compute *structure mappings*—sets of systematic correspondences between the elements of different representational systems—renders a symbol system functional because it makes it possible to use that system to reason about other systems. For example, to use an equation such as *f* = *ma* (a symbolic expression) to solve a physics problem, it is necessary to correctly map the elements of the problem onto the variables *f*, *m*, and *a* (Ross, 1989). It is only when a symbol system (such as an equation) is used to reason about another system that it becomes useful.

But such mappings go beyond arithmetic reasoning. Structure mappings are ubiquitous in human thinking, underlying language comprehension and production (a mapping between linguistic structures and meanings), reasoning based on analogies, schemas, and rules (Faulkenhainer et al., 1989; Gentner, 1983; Gick & Holyoak, 1980, 1983; Holyoak & Thagard, 1989; Hummel & Holyoak, 1997, 2003), explanation (Hummel et al., 2008; Hummel & Landy, 2009), and formal reasoning in domains such as logic, mathematics, and computer programming (Hummel et al., 2014).

It is perhaps in the domain of analogical mapping that the psychological constraints on structure mapping have been most extensively studied, and within this domain researchers have arrived at a broad consensus about the nature of these constraints, which serve to establish structure-sensitive one-to-one mappings between the elements (relations and their arguments) characterizing two domains of knowledge (see Faulkenhainer et al., 1989; Gentner, 1983; Holyoak & Thagard, 1989; Hummel & Holyoak, 1997). We speculate that the same constraints that govern analogical mapping also govern structure mapping in language, deductive logic (Hummel et al., 2014), and explanation (Hummel et al., 2008; Hummel & Landy, 2009).

Without structure mapping, a symbol system would be useless. An empty symbol system disconnected from anything else could follow meaningless instructions, but would not be good for anything else. But armed with structure mapping, a symbolic representation can be used to reason about other structures, including sensory representations. It can make assertions about symbols representing entities in its universe. It can combine the power of symbolic thought and universal generalization (including RRR) to make nontrivial, statistically unjustified inferences about its universe. As a result, it could make inferences outside the scope of its experience, resulting in novel ideas and rapid progress.

**Testing the Hypothesis**

As noted previously, our hypothesis is that hierarchical integration—in the form of both (a) integrating role bindings into multi-place predicates representing relations, and (b) computing and using mappings between two or more such structures—is the minimal difference between an architecture without a symbol system (but which can nonetheless do dynamic binding and therefore make universal inductive generalizations about basic affordances) and human cognition, which is the product of a symbol system.

We hypothesize that both these capacities, together, constitute the minimal requirements for the most basic form of symbolic thought; neither alone is sufficient. An alternative hypothesis is that either multi-place relations *or* structure mapping is sufficient. It is at least logically possible that any task that seems to require multi-place predicates can be solved by a cognitive architecture limited to single-place predicates, as long as that architecture is capable of structure mapping. For example, the LISA model of analogical reasoning (Hummel & Holyoak, 1997, 2003), like people, is capable of solving analogies that violate the "*n*-ary restriction", a constraint on some models of analogy (e.g., Falkenhainer, et al., 1989), which dictates that a predicate with *n* roles can only map to another predicate with *n* roles. LISA, like people, can correctly solve the analogy between *kill* (Bill, Fred) and *cause* (Jack, *die* (Joe)), where the former composes two role bindings and the latter three. Accordingly, it is possible that the multi-place predicate *larger-than* ($x, y$) can be replaced by the single-place predicates *large* ($x$) and *small* ($y$) without any cost in accuracy. Likewise, it is possible that any task that seems to require structure mapping can be solved by a cognitive architecture without structure mapping, provided that architecture is capable of representing multi-place predicates.

These possibilities seem especially plausible given that, to our knowledge, no one has ever investigated the computational capabilities of either one in isolation. It is also plausible given the successes of transformer architectures, which accomplish impressive feats without obviously possessing either of these capacities (e.g., Lake & Baroni, 2018, among many others).

**Four Kinds of Cognitive Architectures**: To investigate these questions, we started with a fully symbolic biologically-inspired neurocomputational system, the LISA model of relational reasoning (Hummel & Holyoak, 1997, 2003; see also Doumas et al, 2008, 2022). We then systematically stripped this system of its capacity for hierarchical integration and observed what kinds of problems it remained capable of solving. Specifically, we left LISA with its capacity for dynamic binding (making it like a cat, monkey, or fruit fly), but stripped it of its ability to (a) combine multiple role bindings into multi-place predicates or (b) compute structure mappings between collections of role-bindings, or both.

We crossed these capacities orthogonally, resulting in four kinds of cognitive architectures, all of which were capable of dynamic binding: (1) The *dynamic binding only* (DBO) architecture is capable of dynamic binding (by synchrony of firing) but neither forming multi-place relations nor structure mapping. (2) The *relations only* (RO) architecture is capable of representing multi-place predicates (i.e., relations) but not structure mapping. (3) The *mapping only* (MO) architecture is capable of structure mapping but not forming multi-place predicates. And (4) the *relations and mapping* (R&M) architecture is capable of both representing multi-place predicates and structure mapping.

In order to keep the resulting architectures on equal footing, none of them possessed the capacity for CWSG. Only the architectures with structure mapping would have been able to use CWSG, so it would have been a confound in our simulation design if we had allowed those architectures to use CWSG as well as mapping. If the absence of either multi-place predicates or system mapping can be compensated by the presence of the other, then the MO and RO architectures should perform similarly. But if the presence of one cannot compensate for the lack of the other, then MO and RO should reveal different patterns of success and failure on the tasks we give them.

A related motivation for these investigations is that LISA, a system that was designed to model complex relational reasoning, has never been tested on tasks that do not require CWSG, multi-place predicates, or system mapping. It remains unclear which of these abilities are required to perform the kinds of tasks investigated in this paper, including tasks that many nonhuman animals can perform. For example, can a chimp's use of tools be understood as an example of structure mapping without multi-place predicates? Can a bumblebee's ability to respond to spatial relations be understood as a case of multi-place predicates without structure mapping? It will also be interesting to observe whether LISA can make symbolic inferences without CWSG, which heretofore has been widely believed to be an essential component of symbolic, or at least analogical, reasoning.

**Four Kinds of Tasks**: We also constructed four kinds of cognitive tasks by orthogonally crossing whether a task requires multi-place predicates, structure mapping, or both. All the tasks were designed to require dynamic binding. Critically, all four tasks were designed to be unsolvable based on featural confounds. That is, the tasks were designed to serve as tests of the kinds of tasks that can be performed by various kinds of cognitive architectures independent of the tasks' semantic content. We are interested in classes of tasks a given cognitive architecture can perform on novel inputs, that is, inputs whose features have not previously been experienced in the context of the task (see also Doumas et al., 2022, for a discussion of the importance of this kind of cross-domain transfer). We are specifically interested in what kinds of generalization various cognitive architectures are capable of, in the most general case. Accordingly, in the simulations reported here, the features of the objects involved in the tasks were completely uninformative about the correct inference. In addition, every task is designed to represent generalization from a single prior example.

**17 Simulations**: As described under Simulations, all four tasks were designed to be perceptual inference tasks akin to perceiving affordances of increasing complexity, according to whether they required representing multi-place predicates, computing structure mappings, or both. All four cognitive architectures were tested on all four kinds of tasks, resulting in a total of 17 different tests/simulations. The reason there are 17 simulations rather than 16 will be explained shortly.

**Transparency and Openness**

The code and other materials for these simulations has been made publicly available in the Open Science Framework repository and can be accessed at https://osf.io/p5zjw. The images produced for illustration purposes in the General Discussion are available at the same link. The simulations presented here were not preregistered.

**The LISA Model**

The starting point for this effort is the LISA model of analogical reasoning (Hummel & Holyoak, 1997, 2003). We chose LISA as our starting point because it is an explainable, biologically-inspired neurocomputational architecture that has been demonstrated to perform tasks that require multi-place predicates and structure mapping. Although there have been several advances in LISA and its successor, DORA (Discovery Of Relations by Analogy; Doumas et al., 2008, 2022) since Hummel & Holyoak's original (1997) publication (e.g., Doumas et al., 2008, 2022; Heaton & Hummel, 2019; Hummel & Holyoak, 2003; Hummel et al., 2008; Hummel et al., 2014; Hummel & Landy, 2009; Knowlton et al., 2012; Licato et al., 2012; Wilner & Hummel, 2017), each of these advances has been directed toward performing increasingly complex reasoning. We chose Hummel and Holyoak's (1997) version of the model because it is both simpler than its successors and adequate to our task, and because its architecture naturally affords removing the capacities for multi-place predicates and mapping. The individual contributions of each of these capacities have never previously been explored.

As such, the description that follows is a description of the Hummel and Holyoak (1997) version of LISA. Almost everything in our description (except for a few details of the mapping algorithm) is true of the subsequent versions of the model. That is, Hummel and Holyoak's

(1997) LISA is mostly a subset of the subsequent incarnations of the model. The details of LISA's operation are described here only in verbal terms. For the details of the model's operation, including all equations, see Hummel and Holyoak (1997), Appendix A.

**Figure 1**
*Illustration of Knowledge Representation in LISA's Long-term Memory ("LISAese")*

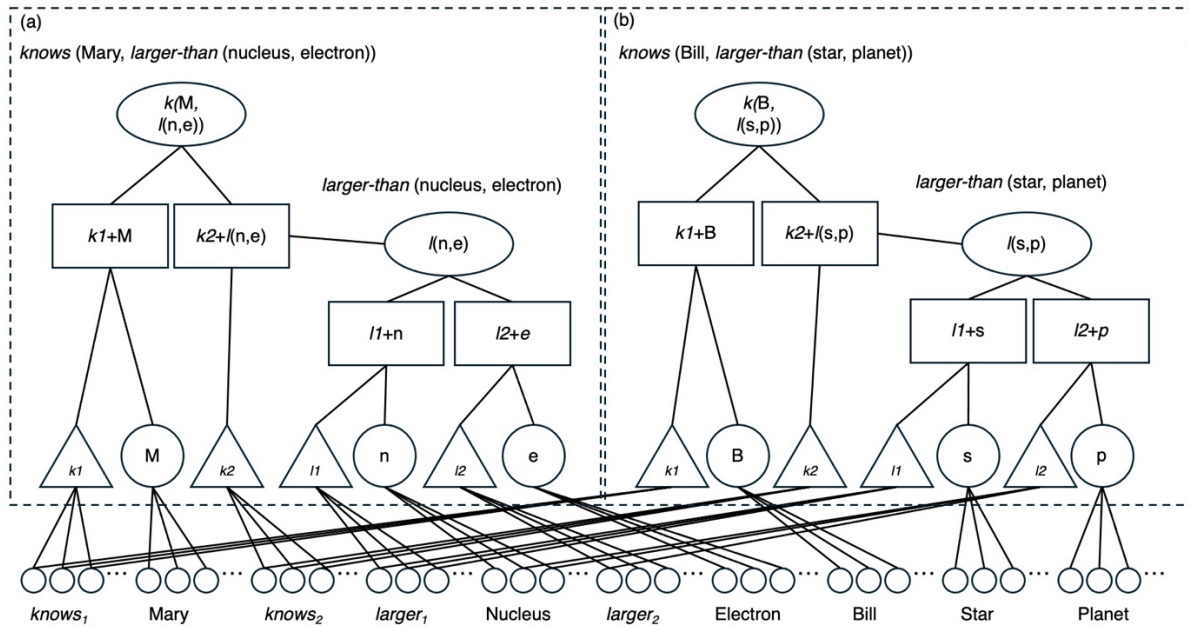

*Note*. Small circles at the bottom of the figure are *semantic* units, large circles are *object* units, triangles are *predicate* units, rectangles are *role-filler binding* units, and ovals are *proposition* units. Thin black lines represent bidirectional excitatory connections between units. Refer to the text for details.

**Knowledge Representation: "LISAese"**: The heart of LISA and its successors is *LISAese*, a basis for representing propositional (i.e., symbolic) knowledge in a distributed fashion. LISA represents propositions using a hierarchy of distributed and progressively more localist codes (Figure 1). At the bottom of the hierarchy, *semantic units* (small circles in Figure 1) represent objects and relational roles in a distributed fashion. Consider, for example, how LISA might represent an analogy between a scientist reasoning about atoms and another scientist reasoning about the solar system. The object "Mary" in the proposition *knows* (Mary, *larger-than* (nucleus, electron)) might be represented by features such as "human", "adult", "female", "scientist", etc., and the *knower* role of the *knows* relation might be represented by features such as "knowledge", "ontology", "belief", etc.[3] At the next level of the hierarchy, objects and predicates (e.g.,

---
[3] One limitation of the 1997 and 2003 versions of LISA is that the semantic representations of objects and predicates must be specified by hand. However, Wilner and Hummel (2017) developed an algorithm for generating these representations automatically, without hand-coding on the part of the modeler. One limitation of the Wilner and Hummel algorithm is that it produces semantic representations that are extremely difficult to understand and impossible to precisely control in terms of their similarity to one another. The simulations reported here are

relational roles) are represented by localist units (large circles and triangles respectively in Figure 1), as are role-filler bindings (rectangles in Figure 1) and complete propositions (ellipses in Figure 1). As illustrated in Figure 1, predicates can take either objects (e.g., Mary) or whole propositions (e.g., *larger-than* (nucleus, electron)) as arguments; that is, LISAese is a recursive symbolic representation. This example uses objects like Mary and the solar system, but LISA is equally capable of reasoning about representations of objects in a scene of the kind that would be delivered by the visual system.

Although Figure 1 depicts units representing objects, role-bindings and propositions with different shapes for clarity, these units are all the same "kind" of simulated neuron (specifically, leaky integrators; see Hummel & Holyoak, 1997, Appendix A) that differ only in their temporal integration properties. Object, predicate, and role-binding units integrate their inputs on a rapid temporal scale (corresponding roughly to oscillations at the gamma or high gamma frequency range, that is, 40 – 80 hz) whereas proposition units integrate their inputs over a longer temporal scale (corresponding roughly to theta or beta, or about 10 hz; see Doumas et al., 2008, 2022; Hummel & Holyoak, 1997, 2003; Knowlton et al., 2012).

Within a single knowledge structure—i.e., collection of neurons representing the propositions composing a single situation, problem, or schema (a.k.a., an "analog" in Hummel & Holyoak, 1997, 2003)—any object predicate, role-binding, or proposition is represented by a single unit in all the propositions in which it occurs. For example, "Mary" would be represented by the same unit in the proposition *knows* (Mary, *larger-than* (nucleus, electron)) as in *knows* (Mary, *orbits* (electron, nucleus)). The same is true of the predicate units representing the roles of the *knows* relation.

However, a given object or predicate is represented by different units in different systems. For example, in a system about Bill's knowledge of planets, the roles of the *knows* and *larger-than* relations would be represented by different predicate units than those representing *knows* and *larger-than* in the system about Mary's knowledge about atoms. In this sense, units representing objects, predicates, role-bindings, and propositions represent *tokens* of these entities in specific systems, so we refer to these units collectively as *token units*. By contrast, the semantic units represent *types*, which are shared by the object and predicate units in all systems. For example, even though the roles of *knows* are represented by separate token units in the Mary and Bill systems, they are likely to be connected to many or all the same semantic (type) units.

Because they are shared by the token units in all systems, the semantic units bootstrap the communication between systems and thus serve as the starting point for all the operations LISA performs, including memory retrieval, analogical mapping (Hummel & Holyoak, 1997), analogical inference, schema induction (Hummel & Holyoak, 2003; Hummel et al., 2014; Wilner & Hummel, 2017), relation discovery (Doumas et al., 2008, 2022), explanation, deduction (Hummel, et al., 2008, 2014), and response/action selection (Castro et al., 2023; Doumas et al., 2022).

Figure 1 illustrates the representation of propositions in LISA's long-term memory (LTM). For this purpose, the bindings of predicates (including relational roles) to their arguments (including objects and complete propositions) and the composition of individual role bindings

---

therefore based on hand-coded semantic representations both for clarity and, more importantly, to precisely control the similarity relations among the objects and predicates (including relational roles) composing the tasks and stimuli used in the current simulations. It is important to note that the names of the units (e.g., "Mary", "female", etc.) have no meaning whatsoever to LISA and are included only to make the simulations understandable to human readers.

into complete propositions are represented conjunctively via the role-binding and proposition units respectively.

When a proposition enters LISA's working memory (WM), these bindings are also represented dynamically by systematic synchrony and asynchrony of firing. For example, when LISA "thinks about" (i.e., activates in WM) the proposition *larger-than* (nucleus, electron), the corresponding proposition unit becomes active and excites the role-binding units representing *larger*+nucleus and *smaller*+electron. These units inhibit one another, and due to random noise in the system, one of them (say, *larger*+nucleus) will initially win the inhibitory competition, becoming active and inhibiting the other (*smaller*+electron), rendering it inactive. Once active, *larger*+nucleus will activate the predicate unit *larger* and the object unit "nucleus", which will activate the semantic units to which they are connected. Role-binding units (aka, "SPs" or "sub-propositions"; Hummel & Holyoak, 1997) are connected to yoked inhibitory units, which become active once the (excitatory) role-binding unit has been active above a threshold for a fixed number of iterations (corresponding to about 12 - 25 ms, i.e., 40-80 Hz). When the inhibitor becomes active, it inhibits the excitatory unit to inactivity, allowing other role-binding units (in this case, *smaller*+electron) to become active. This excitatory/inhibitory exchange causes separate role-bindings under the same proposition, along with their relational roles and the roles' arguments, to oscillate out of synchrony with one another.

The result on the semantic units is a pattern of oscillatory activity in which semantic units representing relational roles fire in synchrony with the semantic units representing the objects bound to those roles and out of synchrony with other role-filler bindings. In the case of recursive structures, in which a proposition takes another proposition as an argument (e.g., as in *knows* (Mary, *p*), where *p* is the proposition *larger-than* (nucleus, electron)), the token unit for the lower-level proposition, *p* (*larger-than* (nucleus, electron)), fires in synchrony with the semantics of the predicate (the semantics for *known-thing*), but the role-bindings under *p* (*larger*+nucleus and *smaller*+electron) are not allowed to become active (see Hummel & Holyoak, 1997)[4].

**Memory Retrieval**: The resulting synchronized and desynchronized patterns of firing on the semantic units serve as a natural basis for memory retrieval. Patterns of semantic activation generated by propositions in one structure naturally activate similar propositions in other structures in LISA's LTM. For example, when *larger-than* (nucleus, electron) becomes active in the Mary system, the semantic patterns representing *larger*+nucleus and *smaller*+electron will tend to overlap with, and therefore activate, the semantic patterns encoded by *larger*+star and *smaller*+planet in the Bill system. These units are connected to the proposition *larger-than* (star, planet) in the Bill system. In this way, when LISA "thinks about" the Mary system, it is likely to be reminded of the Bill system. Integrated over the multiple propositions composing any given system, this simple algorithm provides a remarkably complete account of the literature on analogical retrieval/reminding (Hummel & Holyoak, 1997).

**Structure Mapping**: For the purposes of analogical reasoning, knowledge structures ("analogs") in LISA are divided into two sets: a *driver* and one or more *recipients*. The driver is assumed to be the focus of attention. During analogical reasoning, the driver is initially the structure to be reasoned about: It is the *target* of analogical reasoning, i.e., the structure to be understood in terms of some more familiar *source* (Gick & Holyoak, 1980; Holyoak & Thagard, 1989). During memory retrieval, the recipients are assumed to be all other systems in LTM. The goal of retrieval is to select one structure from LTM to serve as a source—a structure that can be used to understand the target.

---

[4] This convention prevents ambiguous synchrony between the roles of the higher-level proposition (*knows* (Mary, *p*)) and those of the lower-level proposition (*p* = *larger-than* (nucleus, electron)), in this case between the role *known* (belonging to the higher-level proposition) and the role bindings *larger*+nucleus and *smaller*+electron. Allowing *p* to activate its role bindings would result in the patterns *known*+*larger*+nucleus and *known*+*smaller*+electrons on the semantic units, making it unclear which roles applied to "nucleus" and "electrons", or even that *known*+*larger* and *known*+*smaller* were, themselves, concatenations of multiple roles.

Once a potential source has been retrieved from LTM, LISA continues to use the target as the driver, activating propositions one or two at a time, and mapping them onto propositions in the potential source. During the mapping phase of relational reasoning, the source serves as the only recipient, and the mapping from the driver (target) to the recipient (source) is performed by the same algorithm that does retrieval, augmented with a form of Hebbian learning. In effect, LISA simply keeps track of which structures in the driver (target) tend to activate which in the source (recipient), augmented with simple routines that impose a one-to-one mapping constraint (such that any element in one structure is allowed to map to only one element in the other). See Hummel and Holyoak (1997) for additional details.

For our current purposes, the most important aspect of LISA's mapping algorithm is that it is incremental, meaning that it computes mappings between large structures in small pieces. Like human WM, LISA's WM is inherently capacity limited (Hummel & Holyoak, 1997, 2003). Therefore, unlike SME (Faulkenhainer et al., 1989) or ACME (Holyoak & Thagard, 1989), LISA cannot map entire structures onto other structures all-of-a-piece using graph matching (like SME) or massively parallel constraint satisfaction (like ACME). Instead, LISA maps one structure (the driver) onto another (the recipient) one or two propositions at a time, keeping track of the mappings as it goes, and using mappings discovered earlier to constrain the mappings it discovers later (Hummel & Holyoak, 1997). As a result, the mappings LISA discovers early in the mapping process have a substantial effect on the mappings it discovers later, a property LISA shares with humans (Kubose et al., 2002).

**Analogical Inference, Schema Induction, Relation Learning, and Explanation**: Analogical inference, schema induction, relation learning, and the generation of explanations for novel observations are beyond the scope of the current work, but we mention them because they are all within the scope of LISA's abilities. Just as analogical mapping emerges as a natural consequence of LISAese and LISA's algorithm for memory retrieval, so does analogical inference by CWSG emerge as a natural consequence of LISA's mapping algorithm, as does schema induction by intersection discovery (Hummel & Holyoak, 2003). The ability to learn new relational concepts derives from the same mechanisms as schema induction (Doumas et al., 2008, 2022), and the capacity to generate explanations of novel phenomena derives from the same mechanisms with a systematic relaxation of the one-to-one mapping constraint (e.g., Hummel & Landy, 2009).

**Taking Away LISA's Abilities**

Just as analogical inference, schema induction, and other abilities emerge from LISA's algorithm by adding additional routines to its retrieval and mapping algorithms, it is possible to take LISA in the opposite direction, simulating the cognitive abilities of non-symbolic species by systematically taking away LISA's abilities. In the simulations that follow, we will strip LISA of its ability to (a) form multi-place predicates (i.e., relations) and (b) compute structure mappings, and we will investigate the kinds of problems the resulting systems can solve. We will show that at least some versions of this restricted LISA have abilities that plausibly correspond to the known abilities of some nonhuman species.

## Simulations

Recall that we simulated four kinds of cognitive architectures. *Dynamic Binding Only* (DBO) has the capacity to dynamically bind roles to fillers, but no capacity for either structure mapping or representing multi-place relations. *Multi-place Relations Only* (RO) has dynamic binding and the ability to form multi-place relations but no capacity for structure mapping. *Mapping Only* (MO) has the capacity for dynamic binding and structure mapping but no ability to form multi-place relations. *Relations and Mapping* (R&M) has dynamic binding and the capacity for both multi-place relations and structure mapping.

In order to put all the resulting architectures on equal inferential footing, we removed the capacity for CWSG from all of them. These simulations represent the first time LISA's inferential abilities have been tested without CWSG. They also represent the first time either of LISA's more basic (proto-)symbolic capacities have ever been investigated in isolation. The simulations are designed to test our hypothesis about the origins and fundamental components of symbolic thought.

We removed LISA's capacity for structure mapping (i.e., DBO and RO), by setting $\mu$, the rate at which it learns mapping connections, to zero. In the architectures with the capacity for structure mapping (i.e., MO and R&M), we left $\mu$ at its default nonzero value (i.e., 0.9).

We removed LISA's capacity to represent multi-place relations (DBO and MO) by replacing any two-place predicate (i.e. relation) with two single-place predicates whose semantic features were otherwise identical to the corresponding roles of the two-place relation. For example, the relation *larger-than* $(x, y)$ would be replaced by the predicates *large* $(x)$ and *small* $(y)$, where the semantic features attached to *large* () are identical to those on the first role of *larger-than* () and the features attached to *small* () are identical to those on the second role of *larger-than* (). Equating the semantic features of the relations and predicates serves to ensure that the only difference between the multi- and single-place versions of the model is that the multi-place architectures can integrate multiple predicates into a single, hierarchical structure (i.e., a multi-place relation), whereas the single-place architectures cannot; in terms of the semantic features at their disposal, all the architectures are identical.

In general, all the simulations reported here were carefully controlled to ensure that there were no featural differences between the objects and predicates across architectures. This control ensures that no differences in the architectures' performance can be attributed to the featural coding of the objects, predicates, or relations used in the simulations. An important principle of neural modeling is that many networks can learn to exploit any feature that is predictive of the correct response (even a single pixel in an image, as demonstrated by Malhotra et al., 2020). A related principle is that a neural network with an appropriate architecture can perform any task on which it is directly trained. Although LISA is not trained in the conventional machine learning sense, in the simulations we nonetheless controlled the features on the objects in the perceptual structures and LTM so that they were uninformative about the correct answer.

**Simulation Format**

Every simulation was run with two structures ("analogs"), a *Perception* analog, representing a visual stimulus confronting an animal[5], and a *Memory* analog representing a similar situation in the animal's long-term memory (see Figure 2). The Perception analog always served as the driver. Facts (propositions) in Perception became active one at a time, activating facts in Memory, which served as the recipient. To place the various architectures on equal footing, the perceptual affordance was always represented as a single semantic unit, *Affordance*, attached either to one predicate in Memory (in the case of architectures without multi-place relations, i.e., DBO and MO) or to one role of a relation in Memory (in the case of architectures with multi-place relations, i.e., RO and R&M). The Affordance feature was never attached to any object or role in Perception. Instead, each architecture's task was to use the Memory analog to decide whether the affordance (represented by the unit *Affordance*) applied to any object in Perception and therefore ought to be bound to that object. Our reliance on semantic feature binding to represent inferences puts all the architectures on equal footing, in the sense that every kind of architecture can perform this form of inference. As noted previously, none of the simulations require CWSG, which is only possible in the architectures that can compute mappings (MO and R&M). For this reason, even the most complex inferences simulated here (i.e., the fully symbolic R&M ones) are simple compared to the full range of human symbolic thoughts.

---
[5] The Perception analog, like the Memory analog, was represented in LISAese, not as a literal image.

The Perception analog always contained one *Critical* object to which the perceptual affordance applied. For example, if the Critical object is a horizontal surface of a particular size and shape, then the affordance *can-walk-on* applies to that object. A successful inference was operationalized as the model dynamically binding the *Affordance* semantic unit to the Critical object semantic unit (here, the surface) by causing them to fire in synchrony with one another. As a control, the Perception analog also contained one other *Noncritical* object, which should not be bound to the affordance.

**Figure 2**

*LISAese Representation of the Dynamic Binding Only (DBO) Task*

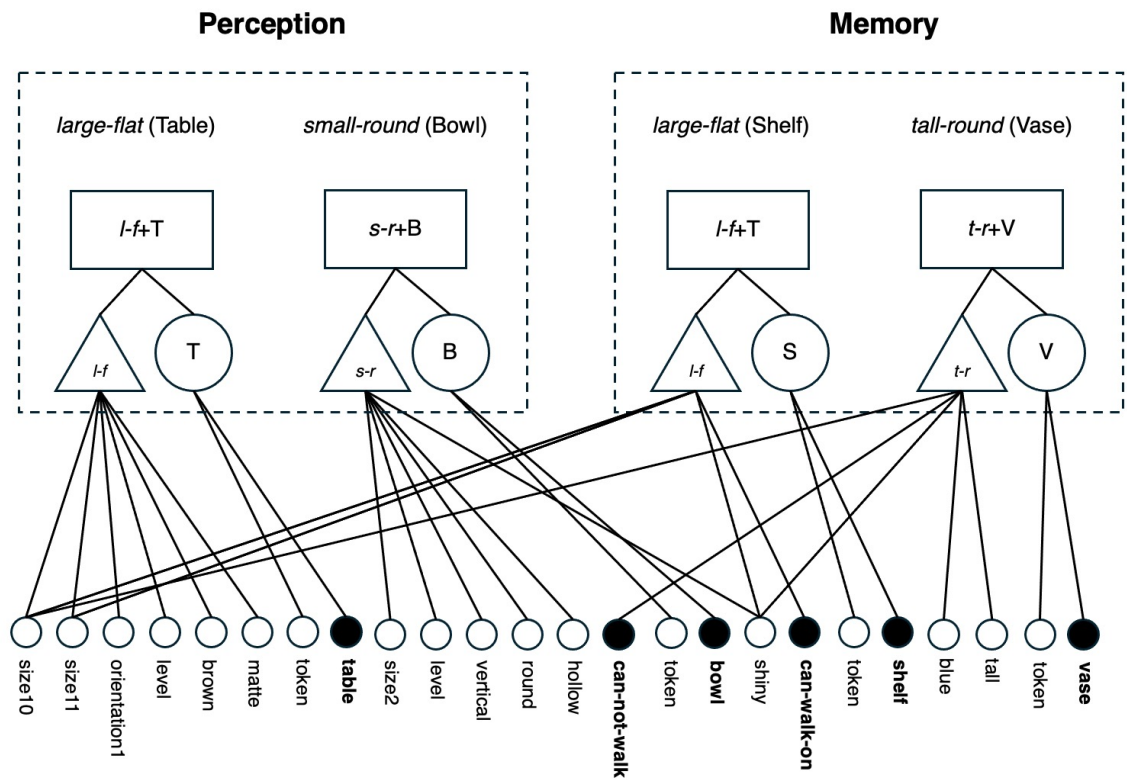

*Note.* Semantic units depicted as filled black circles represent the important objects and properties in the simulation. "table" is the Critical object in Perception, "can-not-walk-on" is the non-affordance, "bowl" is the Other object in Perception, "can-walk-on" is the affordance, "shelf" is the Target object in Memory, and "vase" is the Distractor object in Memory. Electrodes are placed in "table", "bowl", "can-walk-on" and "can-not-walk-on". They are not placed in "vase" or "shelf", but "vase" will tend to fire in synchrony with "can-not-walk-on" and "shelf" will tend to fire in synchrony with "can-walk-on".

      To determine whether any given architecture was able to make the correct inference about the affordance, we placed virtual "electrodes" in the semantic units representing the *Affordance* feature, a *Critical* feature (i.e., a semantic feature attached to the Critical object in Perception), a *No-affordance* feature (attached to a predicate other than the predicate containing the affordance), and a *Noncritical* feature (attached to the Noncritical object in Perception).

      All simulation results are presented as time traces of the activations of these four semantic units over the course of the simulation. A successful inference is one in which the *Critical* semantic unit fires in synchrony with the *Affordance* semantic unit (and out of synchrony with the *No-affordance* semantic unit; see, e.g., Figures 4…7), and a failure of inference is reflected in the *Affordance* semantic unit firing out of synchrony with the *Critical* semantic unit or firing in synchrony with the *Noncritical* semantic unit.

**Task Overview**

      Each of the four model architectures was run on four tasks, each of which required the capacity for dynamic binding, but which varied systematically in their dependence on structure mapping and the capacity to represent multi-place relations: The *Dynamic Binding Only* (DBO) task was designed to require the capacity to dynamically bind an object to a single-place predicate but to not require structure mapping or multi-place relations. We took this task to be equivalent to inferring a basic perceptual affordance. The *Relations Only* (RO) task requires the capacity to represent multi-place relations and dynamically bind them to their arguments but does not require structure mapping. This task is equivalent to reasoning about a relation between two or more objects in the world (e.g., "Can this object fit inside that one?"). The *Mapping Only* (MO) task requires dynamic binding and structure mapping but does not require multi-place relations. This task requires the capacity to keep track of object correspondences across contexts, for example deciding that object *A* corresponds to object *B* based on shape and remembering that correspondence even though *A* and *B* differ in color. The *Relations and Mapping* (R&M) task requires both multi-place relations and structure mapping. This task requires what we regard as the capacity for symbolic thought (see also Penn et al., 2008), but without CWSG.

      The simulations reported here were designed to elucidate broad classes of tasks various kinds of cognitive architectures can and cannot perform in general. Accordingly, in addition to providing concrete examples of all four kinds of tasks, each task is also described in terms of abstract objects, predicates, relations, and semantic features to emphasize the fact that these simulations, and the computational capacities they illustrate, are based entirely on the abstract structure of the task to be performed and the architectures attempting to perform them.

**Dynamic Binding Only Task (DBO): Attributes and World-to-Self Relations**

The DBO task was designed to mimic an animal responding to a basic perceptual affordance, such as a cat deciding whether a surface is suitable for walking. As summarized in Table 1 (see also Figure 2), Perception presented two objects, a table (the Critical object) and a bowl on the table (the Noncritical object). The Critical object was attached to the semantic units *table* (the *Critical* semantic unit) and *token*, and the Noncritical object was attached to *bowl* (the *Noncritical* semantic unit) and *token*. These semantic units represent idiosyncratic features of these objects, such as their color, texture, location in the visual field, etc., which can be used to individuate them as specific tokens (e.g., object files; Kahneman, Treisman, & Gibbs, 1992). Each object in Perception shared an equal number of semantic features with every object in Memory (namely, the unit *token*), making it impossible to solve the task based on the featural similarity of the objects involved, and reflecting our assumption that the perceptual display is novel in the animal's experience. In all the simulations reported here, the relevant perceptual properties of the objects involved were carried, not by semantics connected to the objects themselves, but by semantics attached to the predicates bound to those objects (Figure 2):[6] These predicates, such as *large-flat* () and *small-round* (), represented collections of visual properties (i.e., semantic features), denoted by various semantic units. The names of these units are meaningless to LISA and are included here only to make the simulations easier to understand.

The predicate unit *large-flat* () was connected to semantics that represent various visual properties, some of which were relevant to the affordance (e.g., *can-walk-on*). The predicate *large-flat* () was bound to the Critical object (e.g., the Table) by the proposition *large-flat* (Table). The predicate *small-round* () was connected to semantics that represent its visual properties (e.g., those of a bowl), none of which suggest walkability. *Small-round* () was bound to the Noncritical object (the bowl) by the proposition s*mall-round* (Bowl).

The Memory analog in the DBO simulation represents a situation in which there are again two objects, a Target (e.g., a Shelf), which is bound to a predicate, *large-flat* (), which is connected to the *Affordance* semantic unit (e.g., *can-walk-on*) and a *Distractor* (e.g., a vase), which is bound to a predicate, *small-round* () that is not connected to the *Affordance* unit but instead to the *No-affordance* semantic unit. As in the Perception analog, the objects Target (e.g., Shelf) and Distractor (e.g., Vase) in Memory were connected only to minimal semantics, namely *token* and *target*, in the case of Target, and *token* and *distractor*, in the case of Distractor. Note that these semantics are insufficient to decide whether the Critical object (the Table) in Perception is more like the Target (the Shelf) or the Distractor (the Vase) in Memory, since the Critical object in Perception shares exactly one semantic feature (namely, *token*) with both the Target and Distractor objects in Memory.

This convention is followed in all the simulations reported here: The object semantics never disambiguate the correct inference. However, the predicates in Perception and Memory disambiguate the correspondence. The predicate *large-flat* () in Memory is attached to the semantics *vision1*, *vision2*, *vision3*, *vision4*, *vision7*, and *Affordance* (the semantic unit representing the affordance to be inferred about the Critical object). The predicate *large-flat* () in Memory shares the semantic features *vision1…vision4* with the predicate *large-flat* () in Perception (which is bound to the Critical object), but it only shares *vision7* with the predicate *small-round* (), which is bound to the Noncritical object in Perception. Therefore, the semantic features favor the mapping from

---

[6] The separation of visual features into "object" and "predicate" semantics was not strictly necessary for the DBO simulations, but we adopted this convention to keep these simulations as similar as possible to the other simulations.

*large-flat* () in Perception to *large-flat* () in Memory over the mapping to *small-round* () in Memory. This mapping causes the *Affordance* unit to fire in synchrony with the Critical object's semantic unit, *Critical*.

**Logic of the Simulation**:   The logic of this simulation is as follows: Any cognitive architecture that can map the predicate *large-flat* () (e.g., *large-flat* (table)) in Perception to the predicate *large-flat* () (e.g., *large-flat* (shelf)) in Memory will thereby activate the *Affordance* unit (attached to the *large-flat* () predicate in Memory) at the same time as—thereby binding it to—the Critical object (the Table) in Perception. Such an architecture will have correctly inferred that the *Affordance* feature applies to the Table: The cat will have inferred that it can walk on the table. We therefore expect any architecture capable of dynamically binding a predicate to an argument (i.e., all four architectures tested here) to be able to infer that the affordance applies to the Critical object, regardless of the other properties of that object.

Table 1: The DBO Task: Basic Perceptual Affordance

|  | **Abstract** | | **Example** | |
| --- | --- | --- | --- | --- |
|  | **Perception** | **Memory** | **Perception** | **Memory** |
| **Propositions** | *vision1* (Critical) *vision2* (Other) | *affordance* (Target) *no-afford.* (Dist.) | *large-flat* (Table) *small-round* (Bowl) | *large-flat* (Shelf) *tall-round* (Vase) |
| **Semantic coding of predicates** | *vision1*: [v1, v2, v3, v4, v5, v6]  *vision2*: [v7, v8, v9, v10, v11, v12] | *affordance*: [v1, v2, v3, v4, v7, **affordance***]  *no-afford.*: [v1, v7, v8, v9, **no-affordance***] | *large-flat*: [size10, size11, orn.1, level, brown, matte]  *small-round*: [shiny, size2, level, vertical, round, hollow, can-not-walk*] | *large-flat*: [size10, size11, orn.1, level, shiny, **can-walk-on**]  *tall-round*: [size10, shiny, blue, tall, **can-not-walk-on**] |
| **Semantic coding of objects** | Critical: [token, **critical***]  Other: [token, **other***] | Target: [token, target]  Distractor: [token, distractor] | Table: [token, **table**]  Bowl: [token, **bowl**] | Shelf: [token, shelf]  Vase: [token, vase] |

**Results and Discussion**: As shown in Figure 3, all four architectures succeeded on the DBO (Dynamic Binding Only) task. This result can be seen in the Figure as the *Critical* semantic unit firing in synchrony with the *Affordance* semantic unit. This synchrony of firing indicates that the DBO architecture (and all the others) correctly bound the affordance to the Critical object. This result suggests that DBO tasks, such as perceiving basic affordances, depend only on the ability to (a) represent relevant perceptual properties (such as the size, orientation, and flatness of a surface) explicitly, which is to say independently of one another, and (b) dynamically bind those properties to various arguments (e.g., objects or surfaces) as necessary. As elaborated in the General Discussion, the ability to represent relevant properties explicitly is essential to successful interaction with the world but is only useful in an architecture that can bind those properties together dynamically.

These simulation results demonstrate that the capacity to represent meaningful properties explicitly and bind them together dynamically is sufficient to permit universal inductive generalization of basic perceptual affordances. Note that the representation of the *Critical* object in Perception (the Table) has as no more in common with the *Target* object in Memory (the Shelf) than it does with the *Distractor* (the Vase). They are all simply tokens. But by virtue of the model's ability to dynamically bind the *Critical* (Table) object to its perceptual properties (*large-flat* ()), the model was able to discover that the Table corresponds to the Shelf in Memory and therefore shares with it the perceptual affordance that it can be walked upon. This result helps to explain how animals without the capacity for symbolic thought can nonetheless successfully navigate novel environments.

**Figure 3**
Simulation Results with the Dynamic Binding Only (DBO) Task

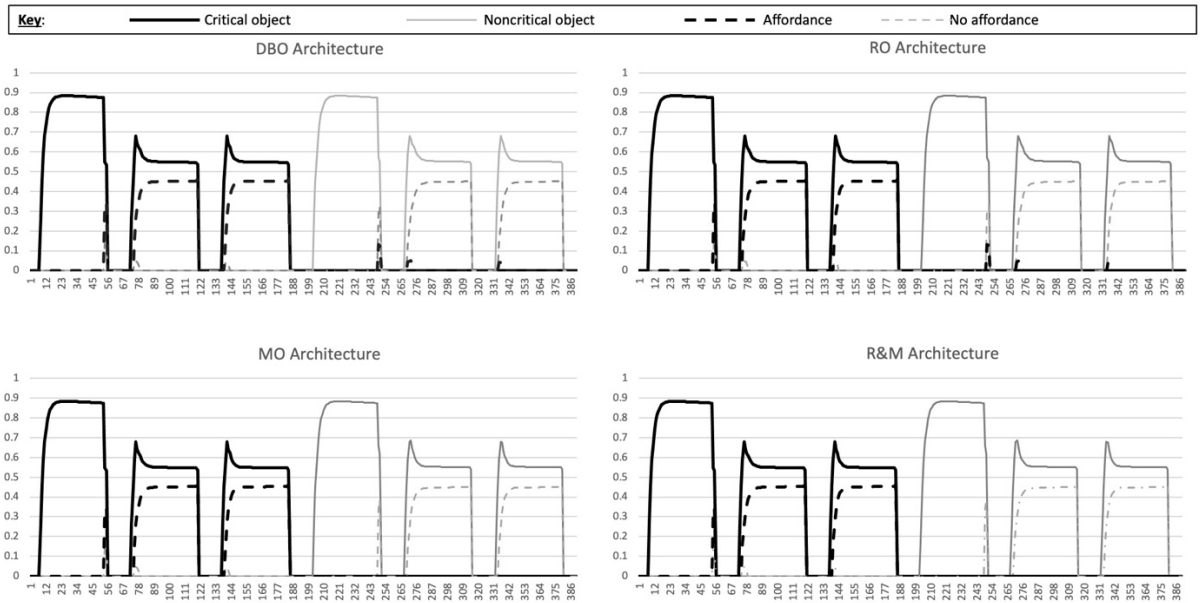

*Note*. Unit activation is on the *y* axis and time (in iterations) on *x*. Solid lines represent the activation of units in Perception; dashed lines are units in Memory. Black lines are the *Critical* object in Perception and the *Affordance* (i.e., the Target object) in Memory. Gray lines are the *Noncritical* ("Other") object in Perception and the *Non-affordance* (the Distractor object) in Memory. A successful inference that the *Affordance* applies to the *Critical* object is represented by the black solid line firing in synchrony (i.e., with activation rising and falling together) with the black dashed line (the *Affordance*). The four graphs represent the four cognitive architectures: DBO (Dynamic Binding Only), RO (Multi-place Relations Only), MO (Mapping Only), and R&M (Multi-place Relations and Mapping). The results demonstrate that all four architectures succeeded in making the correct inference on the DBO task.

Another important result of these simulations is that the more sophisticated architectures performed the task as rapidly as the most basic DBO architecture. As can be seen in Figure 3, all four architectures got the *Affordance* unit firing in synchrony with the *Critical* unit in roughly the same number of iterations. This result suggests that adding the capacity to form multi-place relations and/or the capacity to compute structure mappings need not interfere with a cognitive architecture's ability to perform simpler functions. Finally, as illustrated by this example, architectures without the capacity for structure mapping, such as DBO and RO, can nonetheless compute individual mappings (such as the mapping from the *Critical* object to the *Target*) and exploit them to make simple inferences (e.g., that the *Affordance* applies to the *Critical* object). The difference between such architectures and those that can compute structure mappings (such as MO and R&M) is that the latter, but not the former, can also remember the mappings they have discovered and use them to both constrain future mappings (as in the MO and R&M simulations below; see also Hummel & Holyoak, 1997) and make much more complex inferences using CWSG.

**Relations Only Task (RO): Open-Ended (World-to-World) Relations**

It is one thing to generalize a relation between the world and oneself such as, "I can walk on that surface", a relation that can be compressed into a single-place predicate. But it is a different matter to generalize, or even represent, a relation between multiple objects in the world, where two or more arguments are free to vary, and which therefore cannot be so compressed. Such tasks include, for example, reasoning about whether one object can fit inside another, or whether one object can be placed atop another (see Penn et al., 2008). It is also tempting to speculate that Blaisdell et al.'s (2006) rats, who can respond to causal relations, and Martin-Ordas's (2023) bumblebees, who respond (in limited situations) to spatial relations, may belong in the RO cell of our matrix.

Because we are simulating the absence of multi-place relations in LISA by replacing the roles of a relation such as *smaller-than* () with a collection of single-place predicates that represent the same semantic features as the roles of the relation (e.g., *small* () and *large* ()), it was possible to make two versions of the Dynamic Binding Only (DBO) architecture: The *Cat* version of DBO (Table 2.1) corresponds to a cognitive architecture, such as a cat, that cannot perceive allocentric spatial relations (i.e., relations between objects in the world) and has never been able to, and therefore has no such relations in memory. The *Balint's* version of the DBO architecture (Table 2.2) corresponds to a cognitive architecture, such as a person with Balint's syndrome, who cannot currently perceive allocentric spatial relations, but who once could perceive them, and therefore has them in memory. Balint's syndrome is caused by bilateral damage to the posterior parietal lobes and is characterized by symptoms such as *simultanagnosia*, an inability to perceive more than one object at a time. As a result, people with Balint's syndrome have difficulty perceiving the spatial relations between objects. We placed the Balint's architecture in the DBO cell of the architecture matrix for convenience because the RO task does not require structure mapping; we are not claiming that people with Balint's syndrome cannot compute structure mappings.

As before, the Relations Only (RO) architecture (Table 2.3) corresponds to a cognitive architecture that both can perceive allocentric relations and has memories of such relations but cannot compute structure mappings. The Mapping Only (MO) architecture can compute structure mappings but cannot represent multi-place relations. The Relations and Mapping (R&M) architecture can both represent multi-place relations and compute structure mappings. In all five architectures, the task was to infer that a small toy in perception can be placed atop a large box based on the allocentric multi-place relation *larger-than* ().

In the Cat and Balint's (DBO) architectures, Perception presented two objects, one of which was *small* and the other of which was *large*. In the relational (RO and R&M) architectures, one object (the Toy) was *smaller-than* the other (the Box). In the Balint's and relational versions of the simulation, the Memory analog contained one proposition stating that a Mouse is *smaller-than* a Breadbox (i.e., *smaller-than* (Mouse, Breadbox)). In both these versions of the task, the affordance that the smaller object can be placed atop the larger one was a semantic feature, *Affordance*, of the first (*smaller*) role of the *smaller-than* () relation (see Tables 2.2 and 2.3). In the Cat version of the simulation, the same affordance was a feature of one of the *small* () predicates (the one describing the Mouse) in the Memory analog (Table 2.1).

As in the case of the DBO task simulations, all these simulations were run by setting Perception as the driver, Memory as the recipient, and firing each proposition in Perception once. In the case of the MO (mapping only) architecture, we fired the propositions twice each in order (i.e., first, second, first, second; denoted as "double time" in Figure 4) to give the architecture the opportunity to solve the task (during the second firing of the propositions) based on any mappings it discovered (during the first firing of the propositions). As before, those architectures with the capacity for structure mapping (MO and R&M) had μ (the mapping connection learning rate) set to the default of 0.9, whereas those without that capacity (DBO and RO) had it set to zero.

Table 2.1: Cat Version of the Nonrelational (DBO and MO) Architectures:
No Relations in Either Memory or Perception for the Relations Only (RO) Task

|  | **Abstract** | | **Example** | |
|---|---|---|---|---|
|  | **Perception** | **Memory** | **Perception** | **Memory** |
| **Propositions** | *vision1* (Critical) *vision2* (Other) | *affordance* (Target) *no-afford1* (Dist.1) *no-afford2* (Dist.2) *no-afford3* (Dist.3) | *small* (Toy) *large* (Box) | *small1* (Mouse) *large1* (Breadbox) *small2* (Fork) *large2* (Table) |
| **Semantic coding of predicates** | <u>*vision1*</u>: [v1.1, v2.1, v3.1, v4.1, v5.1, v6.1]<br><br><u>*vision2*</u>: [v1.2, v2.2, v3.2, v4.2, v5.2, v6.2] | <u>*affordance*</u>: [v1.1, v2.1, v3.1, v8.1, **affordance***]<br><br><u>*no-afford1*</u>: [v1.2, v2.2, v3.2, v7.2, v8.2]<br><br><u>*no-afford2*</u>: [v1.1, v2.1, v3.1, v4.1, v8.1, **no-affordance***]<br><br><u>*no-afford3*</u>: [v1.2, v2.2, v3.2, v4.2, v7.2, v8.2] | <u>*small*</u>: [size1, size2, color1, shape1, shape2, shape3]<br><br><u>*large*</u>: [size4, size5, color2, shape6, shape7] | <u>*small1*</u>: [size1, size2, color1, shape4, shape5, **can-go-on**]<br><br><u>*large1*</u>: [size4, size5, color3, shape7, shape8]<br><br><u>*small2*</u>: [size1, size2, shape1, shape2, shape8, **can-not-go-on**]<br><br><u>l*arge2*</u>: [size4, size5, color3, color6, shape7, shape9] |
| **Semantic coding of objects** | <u>Critical</u>: [token, **critical***]<br><br><u>Other</u>: [token, **other***] | <u>Target</u>: [token, target]<br><br><u>Distractor1</u>: [token, distractor1]<br><br><u>Distractor2</u>: [token, distractor2]<br><br><u>Distractor2</u>: [token, distractor3] | <u>Toy</u>: [token, **toy**]<br><br><u>Box</u>: [token, **box**] | <u>Mouse</u>: [thing, location3]<br><br><u>Breadbox</u>: [thing, location4]<br><br><u>Fork</u>: [thing, location5]<br><br><u>Table</u>: [thing, location6] |

Table 2.2: Balint's Version of the Nonrelational (DBO and MO) Architectures: Relations in Memory but Not Perception for the Relations Only (RO) Task

|  | Abstract | | Example | |
|---|---|---|---|---|
|  | **Perception** | **Memory** | **Perception** | **Memory** |
| **Propositions** | *vision1* (Critical) *vision2* (Other) | *affordance* (T1, T2) *no-afford.1* (Dist.1) *no-afford.2* (Dist.2) | *small* (Toy) *large* (Box) | *smaller-than* (mouse, breadbox) *small* (fork) *large* (table) |
| **Semantic coding of predicates** | *vision1*: [v1.1, v2.1, v3.1, v4.1, v5.1, v6.1]<br><br>*vision2*: [v1.2, v2.2, v3.2, v4.2, v5.2, v6.2] | *affordance*: [[v1.1, v2.1, v3.1, **affordance\***], [v1.2, v2.2, v3.2, v7.2, v8.2]]<br><br>*no-afford1*: [v1.1, v2.1, v3.1, v4.1, **no-affordance\***]<br><br>*no-afford2*: [v1.2, v2.2, v3.2, v4.2, v7.2, v8.2] | *small*: [size1, size2, color1, shape1, shape2, shape3]<br><br>*large*: [size4, size5, color2, shape6, shape7] | *smaller-than*: [[v1.1, v2.1, v3.1, **can-go-on**], [v1.2, v2.2, v3.2, v7.2, v8.2]]<br><br>*small*: [v1.1, v2.1, v3.1, v4.1, **cannot-go-on**]<br><br>*large*: [v1.2, v2.2, v3.2, v4.2, v7.2, v8.2] |
| **Semantic coding of objects** | Critical: [token, **critical\***]<br><br>Other: [token, **other\***] | Target: [token, target]<br><br>Distractor1: [token, distractor1]<br><br>Distractor2: [token, distractor2]<br><br>Distractor2: [token, distractor3] | Toy: [token, **toy**]<br><br>Box: [token, **box**] | Mouse: [thing, location3]<br><br>Breadbox: [thing, location4]<br><br>Fork: [thing, location5]<br><br>Table: [thing, location6] |

Table 2.3: Relational (RO and R&M) Architectures:
Relations in Both Memory and Perception for the Relations Only (RO) Task

|  | **Abstract** | | **Example** | |
|  | **Perception** | **Memory** | **Perception** | **Memory** |
| **Propositions** | *visual-input* (Critical, Other) | *affordance* (T1, T2)<br>*no-afford.1* (Dist.1)<br>*no-afford.2* (Dist.2) | *smaller-than* (Toy, Box) | *smaller-than* (mouse, breadbox)<br>*small* (fork)<br>*large* (table) |
| **Semantic coding of predicates** | *visual-input*: [[v1.1, v2.1, v3.1, v4.1, v5.1, v6.1], [v1.2, v2.2, v3.2, v4.2, v5.2, v6.2]]<br><br>(*visual-input* is 3.1's *vision1* and *vision2* integrated into a single relation) | *affordance*: [[v1.1, v2.1, v3.1, **affordance***], [v1.2, v2.2, v3.2, v7.2, v8.2]]<br><br>*no-afford1*: [v1.1, v2.1, v3.1, v4.1, **no-affordance***]<br><br>*no-afford2*: [v1.2, v2.2, v3.2, v4.2, v7.2, v8.2] | *smaller-than*: [[size1, size2, color1, shape1, shape2, shape3], [size4, size5, color2, shape6, shape7]]<br><br>(*smaller-than* is 3.1's *small* and *large* integrated into a single two-p[lace relation) | *smaller-than*: [[v1.1, v2.1, v3.1, **can-go-on**], [v1.2, v2.2, v3.2, v7.2, v8.2]]<br><br>*small*: [v1.1, v2.1, v3.1, v4.1, **can-not-go-on**]<br><br>*large*: [v1.2, v2.2, v3.2, v4.2, v7.2, v8.2] |
| **Semantic coding of objects** | Critical: [token, **critical***]<br><br>Other: [token, **other***] | Target: [token, target]<br><br>Distractor1: [token, distractor1]<br><br>Distractor2: [token, distractor2]<br><br>Distractor2: [token, distractor3] | Toy: [token, **toy**]<br><br>Box [token, **box**] | Mouse: [thing, location3]<br><br>Breadbox: [thing, location4]<br><br>Fork: [thing, location5]<br><br>Table: [thing, location6] |

**Logic of the Simulations**: The only differences between the Cat, Balint's and relational versions of this simulation (Tables 3.1…3.3) concern whether the *smaller-than* relation in Perception (see 2.3) and the *smaller-than* relation in Memory (see Table 2.2) were represented as two-place predicates (i.e., *smaller-than* ()) or as pairs of separate single-place predicates (i.e., *small* () and *large* ()). In the relational architectures, both are represented as relations; in the Balint's version, the relation in Memory (i.e., *smaller-than* ()) is represented as a two-place relation but the pair in Perception (*small* () and *large* ()) is not; and in the Cat version, neither is represented as a multi-place relation. The relational and non-relational versions of these predicates were otherwise identical. For example, the first role of the *smaller-than* () relation has the same semantic features as the single-place predicate *small* () and the second role of *smaller-than* () has the same semantics as *large* ().

The semantic features of the predicates, relational roles, and objects were all configured so that, based only on the semantic overlap between the relational roles and/or single-place predicates, the Critical object in Perception (the Toy) would map to the Distractor object in Memory (a Fork), resulting in an incorrect inference (namely, a failure to bind the *Affordance* unit to the *Critical* unit). But any cognitive architecture that can exploit the hierarchical structure of the multi-place *smaller-than* () relation should map the Critical object (the Toy) in Perception to the Target object (the Mouse) in Memory, binding the *Affordance* semantic unit to the *Critical* semantic unit, resulting in the correct inference that the Toy can be placed atop the Box in Perception.

**Results and Discussion**: As before, an architecture was considered to have made the correct inference if the *Critical* semantic unit (representing the Toy in Perception) fired in synchrony with the *Affordance* semantic unit. As shown in Figure 4, the two architectures capable of representing multi-place relations in Perception (namely, RO and R&M) correctly performed this task, whereas the architectures without multi-place relations in Perception (namely, MO and both the Cat and Balint's versions of DBO) did not. Note from the time traces in Figure 4 that in all architectures, the critical object and the non-affordance would tend to fire together due to semantic overlap, but that in the RO and R&M architectures the constraint imposed by the hierarchical structure (i.e., the shared proposition unit) overcomes the initial semantic bias. The MO architecture did no better firing each proposition twice than it did firing them once, demonstrating that its capacity for structure mapping could not compensate for its inability to represent multi-place relations.

**Figure 4**

Simulation Results with the Relations Only (RO) Task

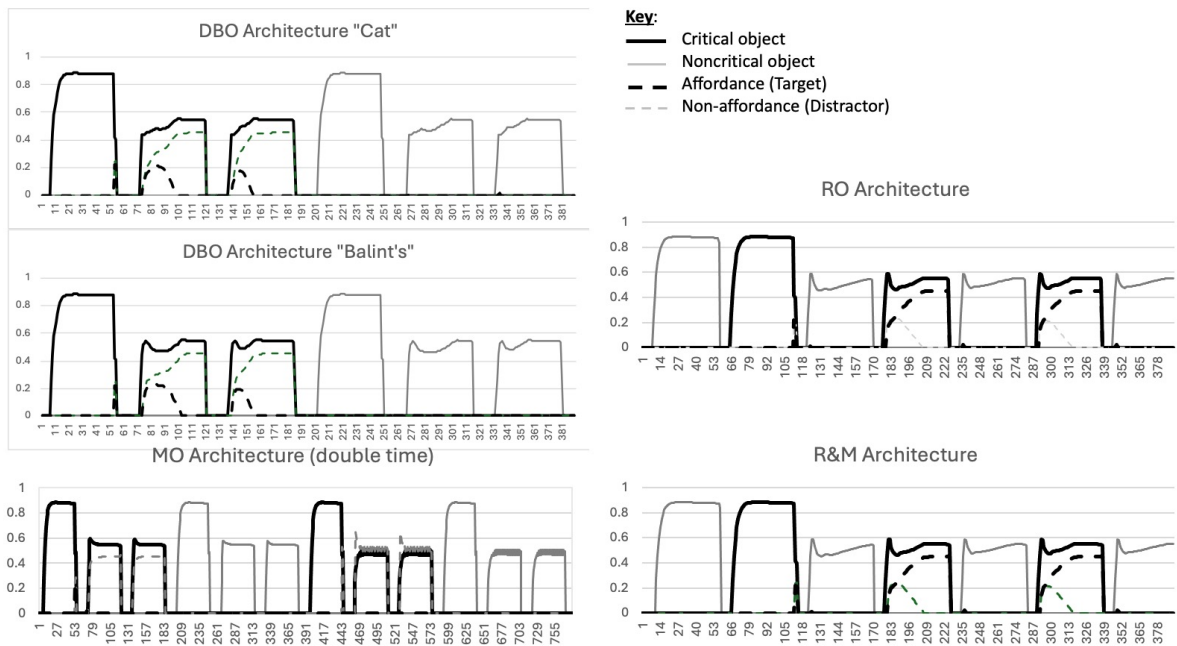

*Note*. The only architectures that succeeded on the RO task (as indicated by the black solid lines rising and falling with the black dashed lines) were those capable of representing multi-place predicates (i.e., relations), namely RO (Relations Only) and R&M (Relations and Mapping). The Mapping Only architecture (MO) failed to solve the task despite being given time to use, in its second consideration of the stimulus, whatever mappings it had learned during the first (the "double time" notation). Note that in all architectures, the critical object and the non-affordance would tend to fire together due to semantic overlap, but that in the RO and R&M architectures the constraint imposed by the hierarchical structure (i.e., the shared proposition unit) overcomes the semantic bias.

Although it may seem obvious that only a cognitive architecture with the capacity to represent multi-place relations can succeed on a task requiring sensitivity to such relations, two aspects of these results are noteworthy. The first is that the Balint's version of the DBO architecture failed, even though this version has the relevant relation in its long-term memory. This result suggests that knowing what a relation is cannot compensate for a cognitive architecture's inability to perceive it. Visual reasoning is an interplay between perception and memory, and unless both have relations in their vocabulary, the architecture is likely to fail on any task that requires sensitivity to such relations. The converse case—the capacity to perceive relations without having any in memory—can be observed in cognitive development when children are first discovering relational concepts. Relational reasoning develops gradually as relations appear, making their way into memory, and we observe a gradual transition from featural to more relational ways of thinking in cognitive development (see Doumas et al., 2008, for a review). The failure of the Balint's version of the simulation also constitutes a novel prediction: Patients with Balint's syndrome should be unable to decide whether it is possible to, say, stack one novel object on top of another.

The second interesting aspect of these results is the failure of the MO architecture. This result suggests that having the capacity to compute structure mappings cannot make up for the absence of the capacity to represent multi-place relations as explicit, hierarchical structures. In turn, this result is consistent with our hypothesis that the two facets of hierarchical integration, structure mapping and multi-place relations, are indeed independent capacities of the human cognitive architecture. As elaborated in the General Discussion, this independence may help to explain why symbolic thought appeared so late in our evolutionary history. Inasmuch as both are critical for symbolic thought and yet independent of one another, it may be that the evolution of symbolic thought was slowed by the need for both mutations to occur in the same organism.

**Mapping Only Task (MO): Memory for Correspondences**

One task that requires the capacity for structure mapping but does not require the capacity for multi-place relations is Marcus's (1998) Identity Function, $f(x) = x$, which requires an architecture to produce as output whatever it is given as an input. Hummel and Holyoak (2003) demonstrated that LISA can solve Marcus's task with mapping and without multi-place predicates. However, their simulation used the full-capacity LISA, complete with CWSG. It remains to be seen whether an analogous task can be accomplished without CWSG. Another such task may be Halford and colleagues' Relational Schema Induction task, in which operators serve as single place predicates and peoples' names serve as their arguments (Halford, et al. 1998; Halford & Busby, 2007).

Another such task, illustrated in Table 3, is analogous to a chimpanzee's use of a stick to retrieve termites from a mound. In this task the Critical object in Perception corresponds to the Target object in memory based on one set of properties, and that correspondence dictates that the Critical object should inherit the Target object's affordance despite mismatching the Critical object on other properties. For example, let the Critical object in Perception be a stick with a particular length and width (represented as the predicate *vision1* in Perception), and the Target object in memory be a stick with similar dimensions (*memory1* in Memory). The *Affordance*, attached to predicate *memory2* in Memory, is that the Target object (and by analogy, the Critical object) can be used to get termites out of a mound. However, there is a Distractor object in Memory whose color and texture (*memory2*) are like those of the Critical object (*vision2* in Perception). If the cognitive architecture can map the Critical object to the Target based on the relevant properties of length and width and remember the mapping to map *vision2* (*Critical*) to *memory3* (*Target*) thereby inferring the affordance ("I can use this to get termites") in spite of the distracting color/texture-similarity between the Critical object and the Distractor, then such an inference would be an example of a structure mapping supporting an inference in the absence of multi-place relations.

Because this simulation does not require us to make accommodations for architectures that cannot represent multi-place relations, the same simulation was run on all four architectures.

**Logic of the Simulation**: Based on the similarity of the shape of the stick (the Critical object) in Perception (*vision1*) to that of the Target object in Memory (*memory1*), the proposition *vision1* (Critical) in Perception will tend to activate the proposition *memory1* (Target) in Memory. A cognitive architecture that can learn structure mappings would therefore learn that the Critical object corresponds to the Target object. Based on the similarity of the color and texture (*vision2*) of the Critical object in Perception to those of the Distractor object in Memory (*memory2*), the proposition *vision2* (Critical) in *Perception* will tend to activate *memory2* (Distractor) in Memory. In this case, *memory2* (which is connected to the semantic unit *No-affordance*) will activate *No-affordance* and the system will fail to infer that the affordance applies to the Critical object. However, to a cognitive architecture that can learn structure mappings, the Critical object is already known to correspond to the Target object in Memory (based on the prior mapping of *vision1* (Critical) to *memory1* (Target)). Given the 1:1 mapping constraint on structure mapping, this prior mapping will tend to favor the mapping of *vision2* (Critical) to *memory3* (Target). In this case, *memory3* will activate the semantic unit *Affordance*, which will fire in synchrony with *memory3* and therefore with the Target (which is bound to *memory3*) and the Critical object's semantic unit, *Critical*. The system will have correctly inferred that the affordance applies to the Critical object.

As before, the simulation was run with Perception as the driver and Memory as the recipient, firing each proposition in Perception once.

Table 3: The Mapping Only (MO) Task: Object/Predicate Correspondences

|  | **Abstract** | | **Example** | |
|---|---|---|---|---|
|  | **Perception** | **Memory** | **Perception** | **Memory** |
| **Propositions** | *vision1* (Critical) <br> *vision2* (Critical) <br> *vision3* (Other) | *memory1* (Target) <br> *memory2* (Distractor) <br> *memory3* (Target) | *long-thin* (Stick1) <br> *color1* (Stick1) <br> *color2* (Stick2) | *long-thin* (Stick3) <br> *color1* (Stick4) <br> *color2* (Stick3) |
| **Semantic coding of predicates** | <u>*vision1*</u>: [v1, v2, v3, v4, v5, v6] <br><br> <u>*vision2*</u>: [v7, v8, v9, v10, v11, v12] <br><br> <u>*vision3*</u>: [v13, v14, v15, v16, v17, v18] | <u>*memory1*</u>: [v1, v2, v3, v4, v19, v20] <br><br> <u>*memory2*</u>: [v7, v8, v9, v10, v21, **no-affordance\***] <br><br> <u>*memory3*</u>: [v13, v14, v15, v16, v22, **affordance\***] | <u>*long-thin*</u>: [v1, v2, v3, v4, v5, v6] <br><br> <u>*color1*</u>: [v7, v8, v9, v10, v11, v12] <br><br> <u>*color2*</u>: [v13, v14, v15, v16, v17, v18] | <u>*long-thin*</u>: [v1, v2, v3, v4, v19, v20] <br><br> <u>*color1*</u>: [v7, v8, v9, v10, v21, **not-termite**] <br><br> <u>*color2*</u>: [v13, v14, v15, v16, v22, **termite**] |
| **Semantic coding of objects** | <u>Critical</u>: [token, **critical\***] <br><br> <u>Other</u>: [token, **other\***] | <u>Target</u>: [token, target] <br><br> <u>Distractor</u>: [token, distractor] | <u>Stick1</u>: [token, **stick1**] <br><br> <u>Stick2</u>: [token, **stick2**] | <u>Stick3</u>: [token, stick3] <br><br> <u>Stick4</u>: [token, stick4] |

**Figure 5**
Simulation Results with the Mapping Only (MO) Task

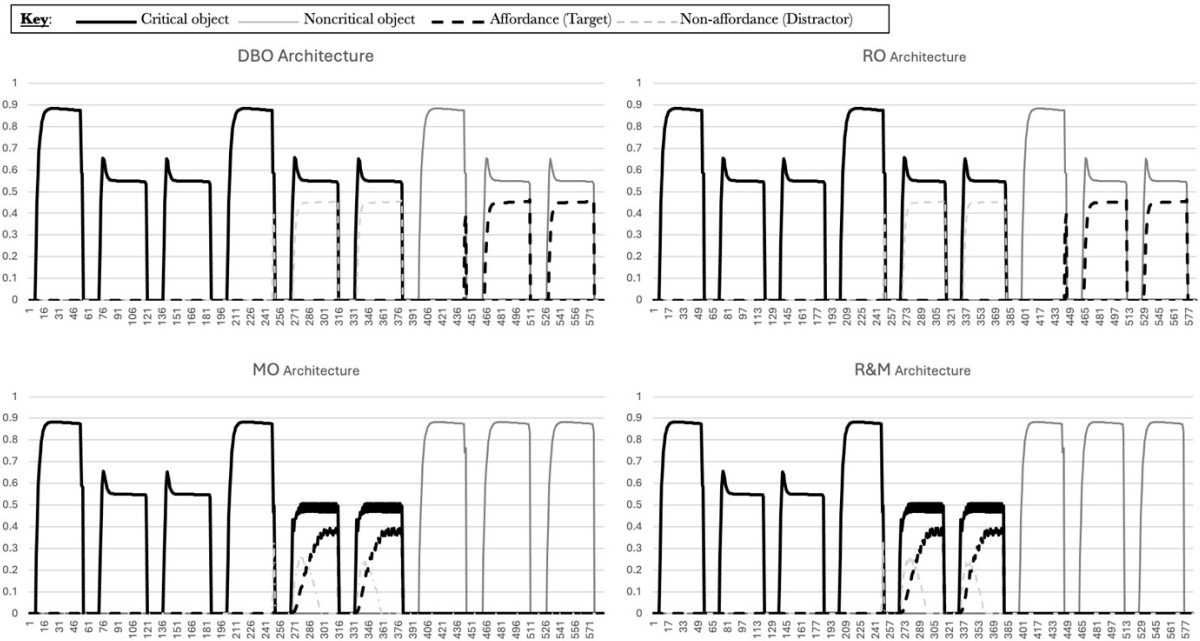

*Note.* The only architectures that succeeded on the MO task (as indicated by the black solid lines rising and falling with the black dashed lines) were those capable of structure mapping, namely MO (Mapping Only) and R&M (Relations and Mapping). The Relations Only architecture (RO) failed to solve the task. Note that prior to learning the mapping connections (i.e., before about iteration 270), both the MO and R&M architectures fail to bind the affordance to the critical object. But after the mappings are learned (i.e., between iterations 270 and 380) they bias the system to activate the affordance along with the critical object. Also note that in all architectures, the critical object and the non-affordance would tend to fire together due to semantic overlap, but that in the MO and R&M architectures the learned mapping connections overcome this semantic bias.

**Results and Discussion**: As shown in Figure 5, the only architectures that solved this task were those capable of structure mapping, namely MO (mapping only) and R&M (relations and mapping). The others were distracted by the color and texture similarity between the Other object in Perception and the Distractor in Memory, as embodied in the semantic overlap between the predicates *vision3* (in Perception) and *memory3* (aka *color2*) in Memory. Note that prior to learning the mapping connections (i.e., before about iteration 270), both the MO and R&M architectures fail to bind the affordance to the critical object. But after the mappings are learned (i.e., between iterations 270 and 380) they bias the system to activate the affordance along with the critical object. Also note in the time trace in Figure 5 that in all architectures, the critical object and the non-affordance start out with a tendency to fire together due to semantic overlap, but that in the MO and R&M architectures the learned mapping connections overcome this semantic bias.

Recall that the RO task proved unsolvable to the MO architecture, demonstrating that a capacity for structure mapping cannot compensate for an inability to represent multi-place relations. In this simulation, the failure of the RO architecture on the MO task demonstrates that a capacity for multi-place relations likewise cannot compensate for an inability to compute and remember mappings between the elements of analogous structures. Together, these results strongly suggest that structure mapping and multi-place relations, although both are kinds of hierarchical integration, are not interoperable or redundant with one another. Instead, they are independent cognitive capacities that are both necessary for basic symbolic thought.

**Relations and Mapping Task (R&M): Symbol Systems**

The final task on which we tested the four architectures was designed to require both structure mapping and multi-place relations. Although we have argued that both these capacities are necessary for basic symbolic thought, the simulations presented here are not intended, or indeed able, to speak to the question of whether they are also sufficient for more complex forms of symbolic thought. The reason is that even this task does not require CWSG or intersection discovery, the algorithms responsible for analogical inference and schema induction in LISA (Hummel & Holyoak, 2003) and relation discovery in DORA (Doumas et al., 2008, 2022). We intentionally designed the R&M task to not require these capacities so that we could use the same metric of performance, namely the ability to dynamically bind the affordance to the Critical object in Perception, for all four architectures. This metric, because it is within reach of the DBO, RO, and MO architectures, is necessarily also weaker than the inferential power of the most sophisticated forms of symbolic thought.

For our current purposes, the R&M task was designed to require multi-place relations and structure mapping without requiring CWSG. One task satisfying these requirements would be operating a new model of a coffee maker (see Tables 4.1 and 4.2). A coffee maker consists of multiple parts in systematic relations that are important to their function (e.g., it is important for the filter basket to be above the carafe) (Figure 6). The inference we designed for the R&M task can be performed without CWSG because it does not require the architecture to infer propositions. Instead, the target inference—that the water tank affords pouring water into—is representing by attaching the semantic *Affordance* (e.g., *can-pour-water*) to the predicate modifying the tank.[7] As with the RO task, we had to design different versions of the R&M task to accommodate the representational limitations of those architectures without the capacity to represent multi-place relations (namely, DBO and MO). Because the RO task simulations revealed that the "Balint's" version of the task (in which relations were resident in Memory but not Perception) performed identically to the "Cat" version (in which neither Perception nor Memory had relations), for the R&M task simulations, we ran only the "Cat" version. The simulations proceeded by firing each proposition in Perception once using Memory as the recipient.

---

[7] Obviously, this is only one of numerous affordances people exploit when making coffee. Others include putting coffee in the filter basket, turning on the switch, grabbing the handle of the carafe to pour the coffee, etc.

**Figure 6**
*Graphical Representation of Reasoning About a Novel Coffee Maker (the R&M Task)*

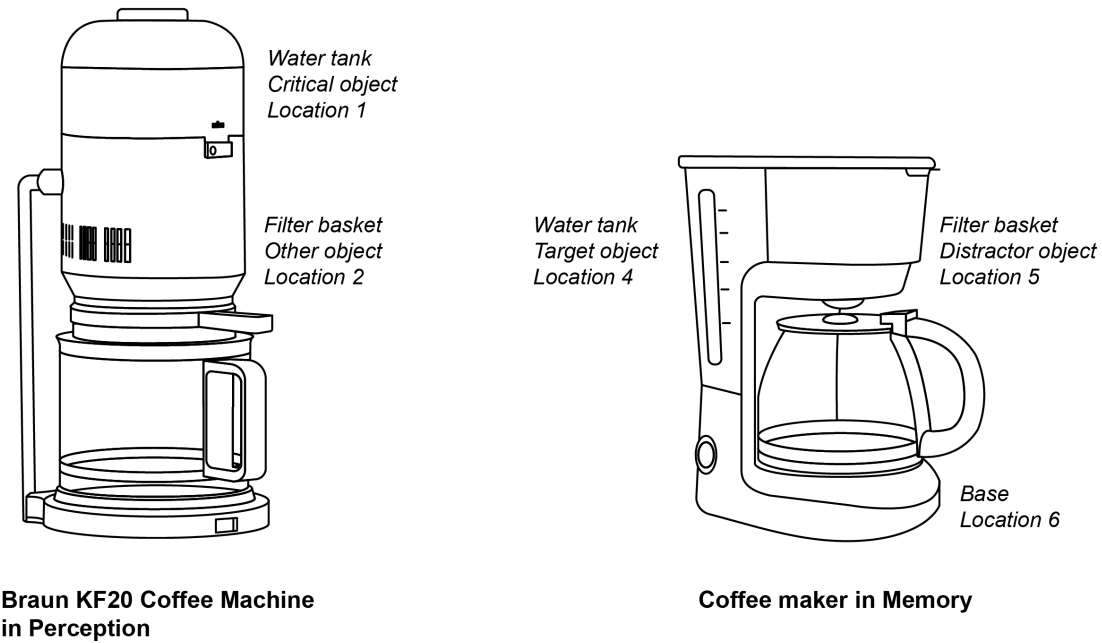

**Braun KF20 Coffee Machine in Perception**

**Coffee maker in Memory**

*Note.* The Braun KF20 Coffee Machine, designed by Florian Seiffert, is an unusual design for a coffee maker. Based on the mapping of the *top-connected* (water-tank, filter-basket) relation in Perception to the *side-connected* (water-tank, filter-basket) in Memory, it is possible to infer that the affordance associated with the coffee maker in Memory also applies to the novel KF20 coffee maker in Perception (namely, that water can be poured into the water tank to make coffee).

**Logic of the Simulation**: The logic of the simulation is easier to follow if we start with the relational version in Table 4.2. The first proposition in Perception states that the water tank of a novel coffee maker is above and connected to the filter basket (*top-connected* (Water-tank, Filter-basket)). In these simulations, the Water-tank in Perception is the Critical object and the Filter-basket is the Noncritical object; in Memory, the Water-tank is the Target and the Filter-basket is the Distractor. Making the correct inference that the Water-tank affords pouring water into requires a relational architecture to map *top-connected* (Water-tank, Filter-basket) in Perception to *side-connected* (Water-tank, Filter-basket) in Memory. To ensure that this mapping is a result of the hierarchical nature of the *top-* and *side-connected* () relations, rather than a trivial consequence of the featural similarity between *top-connected* () in Perception and *side-connected* () in Memory, the first role of *top-connected* () in Perception had more semantic features in common with *location6* (the location of the base of the coffee maker) in Memory than it did with the first role of *side-connected* () in Memory. Likewise, the second role of *top-connected* in Perception had more in common with *open* (a property of the water tank) in Memory than it did with the second role of *side-connected* (four semantics vs. three, respectively). In other words, the semantic features of *top-* and *side-connected* relations and the corresponding locations in the non-relational architectures favor mapping these predicates in Perception to the incorrect predicates in Memory. Therefore, the only way to discover the correct mapping is by using the hierarchical structure of the roles of the *top-* and *side-connected* predicates.

The only difference between the R&M task with relations (Table 4.2) and the version without relations (Table 4.1) is that the *top-* and *side-connected* relations in the relational version were replaced with the single-place predicates *location1* (location of the Water-tank) and *location2* (location of the Filter-basket) in Perception and with *location4* (location of the Water-tank) and *location5* (location of the Filter-basket) in Memory. The two structures are identical in every other respect (compare Tables 4.1 and 4.2). Critically, in Perception, the first and second roles of *connected-to* in the relational architectures (RO and R&M) have identical semantics to the predicates *location1* and *location2*, respectively, in the nonrelational architectures (DBO and MO). In Memory the first and second roles of *connected-to* have identical semantics to the predicates *location4* and *location5*, respectively.

Because finding the correct mapping depends on the hierarchical structure of the *connected-to* relation (and not on the semantic features of their roles), we expect the relational architectures to succeed in mapping *connected-to* (Water-tank, Filter-basket) in Perception to *connected-to* (Water-tank, Filter-basket) in Memory, but we expect the nonrelational architectures to map *location1* (Water-tank) and *location2* (Filter-basket) in Perception to *location6* (Base) and *open* (Carafe), respectively, in Memory. This logic is identical to the logic used in the Relational and "Balint's" versions of the RO simulations, above.

Mapping *connected-to* (Water-tank, Filter-basket) in Perception to *connected-to* (Water-tank, Filter-basket) in Memory also maps Water-tank (the Critical object in Perception) to Water-tank (the Target in Memory). Having established these mappings, a cognitive architecture that can remember and use them will then preferentially map the proposition *open* (Water-tank) in Perception to the proposition *pour-water* (Water-tank) in Memory. By contrast, systems that cannot learn and exploit mappings (like DBO and RO), and systems that cannot represent and use hierarchical multi-place relations (like MO and DBO), will tend to map *open* (Water-tank) to *open* (Carafe) based on the semantic similarity between *open* in Perception and *open* in Memory, thereby failing to infer the affordance about the Target object (the Water-tank) in Perception.

Although we expect the RO architecture to map Water-tank in Perception onto Water-tank in Memory, because it cannot learn this mapping, it will be unable to use it to cause *open* (Water-tank) in Perception to map to the affordance *pour-water* (Water-tank) in memory. In other words, the MO architecture, which cannot represent higher-order relations, will be unable to map the Water-tank in Perception onto the Water-tank in Memory, so it will fail to infer the affordance. By contrast, the RO architecture will be able to map the Water-tank in Perception to the Water-tank in Memory, but it will be unable to use this mapping and will therefore also fail to infer the affordance. As such, we expect only the R&M architecture, which can both represent multi-place relations and learn structure mappings, to successfully infer the affordance on this task.

Table 4.1: R&M Task for Architectures Without Multi-place Relations (DBO and MO)

| | Abstract | | Example | |
|---|---|---|---|---|
| | **Perception** | **Memory** | **Perception** | **Memory** |
| **Propositions** | *vision1.1* (Critical) *vision1.2* (Other) *vision2* (Critical) *vision3* (Other) | *memory1.1* (Target) *memory1.2* (Dist1) *memory2* (Dist2) *memory3* (Dist3) *memory4* (Dist4) *memory5* (Target) | *location1* (water-tank) *location2* (filter-basket) *open* (water-tank) *conical* (filter-basket) | *location4* (water-tank) *location5* (filter-basket) *location6* (base) *open* (carafe) *location7* (on/off-switch) *pour-water* (water-tank) |
| **Semantic coding of predicates** | *vision1.1*: [v1.1, v2.1, v3.1, v4.1, v5.1, v6.1]  *vision1.2*: [v1.2, v2.2, v3.2, v4.2, v5.2, v6.2]  *vision2*: [v7, v8, v9, v10, v11, v12]  *vision3*: [v13, v14, v15, v16, v17, v18] | *memory1.1*: [v1.1, v2.1, v3.1, v19, v20, v21]  *memory1.2*: [v2.1, v2.2, v3.2, v22, v23, v24]  *memory2*: [v1.1, v2.1, v3.1, v4.1, v25, v26]  *memory3*: [v1.2, v2.2, v3.2, v4.2, v27, v28]  *memory4*: [v7, v8, v9, v17, v18, **no-affordance***]  *memory5*: [v11, v12, v13, v14, v15, **affordance***] | *location1*: [v1.1, v2.1, v3.1, v4.1, v5.1, v6.1]  *location2*: [v1.2, v2.2, v3.2, v4.2, v5.2, v6.2]  *open*: [v7, v8, v9, v10, v11, v12]  *conical*: [v13, v14, v15, v16, v17, v18] | *location4*: [v1.1, v2.1, v3.1, v19, v20, v21]  *location5*: [v2.1, v2.2, v3.2, v22, v23, v24]  *location6*: [v1.1, v2.1, v3.1, v4.1, v25, v26]  *open*: [v1.2, v2.2, v3.2, v4.2, v27, v28]  *location7*: [v7, v8, v9, v17, v18, **no-affordance**]  *pour-water*: [v11, v12, v13, v14, v15, **affordance**] |
| **Semantic coding of objects** | Critical: [token, **critical***]  Other: [token, **other***] | Target: [token, target]  Distractor1: [token, distractor1]  Distractor2: [token, distractor2]  Distractor3: [token, distractor3]  Distractor4: [token, distractor4] | Water-Tank: [token, **part1**]  Filter-Basket: [token, **part2**] | Water-Tank: [token, tank]  Filter-Basket: [token, basket]  Base: [token, base]  Carafe: [token, carafe]  On/Off Switch: [token, switch] |

Table 4.2: R&M Task for Architectures With Multi-place Relations (RO and R&M)

|  | Abstract | | Example | |
|---|---|---|---|---|
|  | **Perception** | **Memory** | **Perception** | **Memory** |
| **Propositions** | *vision1* (Crit., Other) *vision2* (Critical) *vision3* (Other) | *memory1* (Target, Dist1) *memory2* (Dist2) *memory3* (Dist3) *mem4* (Dist4) *mem5* (Target) | *top-connected* (water-tank, filter-basket) *open* (water-tank) *conical* (filter-basket) | *side-connected* (water-tank, filter-basket) *location2* (base) *open* (carafe) *location6* (on/off-switch) *pour-water* (water-tank) |
| **Semantic coding of predicates** | <u>*vision1*</u>: [[v1.1, v2.1, v3.1, v4.1, v5.1, v6.1], [v1.2, v2.2, v3.2, v4.2, v5.2, v6.2]]  (*vision1* is 5.1's *vision1.1* and *vision1.2* integrated into a single relation. *vision2* and *vision3* are the same as in 5.1) | <u>*memory1*</u>: [[v1.1, v2.1, v3.1, v19, v20, v21], [v1.2, v2.2, v3.2, v22, v23, v24]]  (*memory1* is 5.1's *memory1.1* and *memory1.2* integrated into a single relation. *memory2*…*mem5* are the same as in 5.1) | <u>*top-connected*</u>: [[v1.1, v2.1, v3.1, v4.1, v5.1, v6.1], [v1.2, v2.2, v3.2, v4.2, v5.2, v6.2]]  (*connected-to* is 5.1's location1 and location2 integrated into a single relation. All other predicates are the same as in 5.1) | <u>*side-connected*</u>: [[v1.1, v2.1, v3.1, v19, v20, v21], [v2.1, v2.2, v3.2, v22, v23, v24]]  (*connected-to* is 5.1's location3 and location1 integrated into a single relation. All other predicates are the same as in 5.1) |
| **Semantic coding of objects** | Same as 5.1 | Same as 5.1 | Same as 5.1 | Same as 5.1 |

**Results and Discussion**: The results, shown in Figure 8, were exactly as expected. Only the R&M architecture correctly solved this task. This result is important because, in combination with the results of the RO and MO simulations, it serves as further evidence that both mapping and multi-place predicates are necessary to solve some kinds of cognitive tasks (such as analogy) and that although both can be viewed as a kind of hierarchical integration of knowledge representations, neither can compensate for the absence of the other.

**Figure 7**
Simulation Results with the Relations and Mapping (R&M) Task

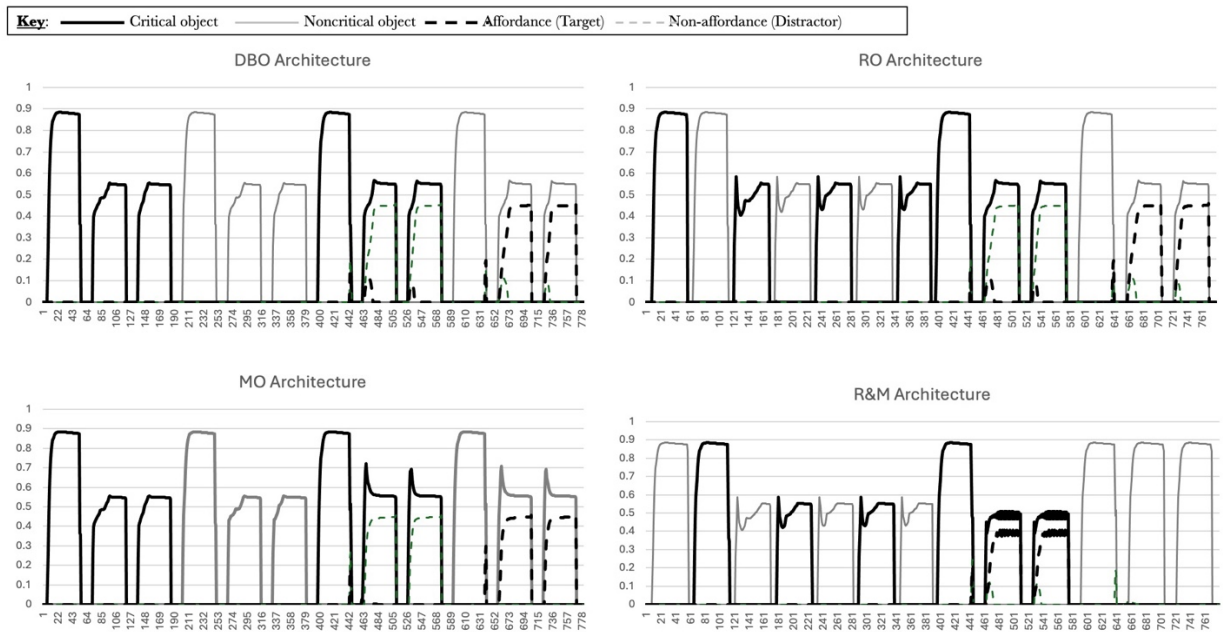

*Note.* Only R&M, the architecture capable of both representing multi-place and computing mappings between structures composed of multiple such relations succeeded in making the correct inference on the R&M ("how to use a novel coffee maker") task.

**General Discussion**

Our goal in this work was to understand the biological underpinnings of symbolic thought. Dynamic binding is a prerequisite to symbolic representation, but many species, including cats, monkeys, and even fruit flies, are capable of dynamic binding, whereas far fewer are capable of symbolic thought. The question we sought to answer is what, in addition to dynamic binding, is required for symbolic thought?

The hypothesis we tested is that two additional capabilities are required to achieve the most basic form of symbolic thought. The first is the capacity to bind multiple role-filler bindings into multi-place predicates expressing relations (e.g., the ability to bind the bindings "*larger* + Box" and "*smaller* + Toy" into the relation, *larger-than* (Box, Toy)). The second is the capacity to map collections of such relations onto one another. A symbol system (such as a language or a system of equations) that cannot be mapped onto specific relevant circumstances (e.g., a situation described by a linguistic utterance, or a problem that can be solved with the system of equations) is useless. Structure mapping is a prerequisite to using any symbol system.

At the outset of this work it was unclear whether multi-place predicates could compensate for the absence of structure mapping and vice versa (i.e., it was unclear whether MO and RO would perform equivalently). In a systematic series of 17 simulations, we tested our hypothesis by investigating which kinds of cognitive architectures could solve what kinds of problems. The results were consistent with our hypothesis. The simulations demonstrated that cognitive architectures capable of representing multi-place predicates can solve problems that architectures without this capacity cannot, and they demonstrated that architectures with the capacity for structure mapping can solve problems that architectures without this capacity cannot. And these simulations demonstrate that architectures with both these capacities are capable of basic symbolic thought.

Our simulation results also revealed several unexpected findings. The first is that adding the capacity to represent multi-place predicates and the capacity to compute structure mappings to a more basic architecture capable of dynamic binding does not slow down or decrease the accuracy of the architecture's ability to perform tasks that do not require those capacities. Recall that all four architectures performed the most basic DBO (dynamic binding only) task with equal speed and accuracy.

The second surprising finding was that performing a task that requires the cognitive architecture to perceive allocentric relations between objects in the world is not facilitated by having representations of relations in memory, unless that architecture can also perceive the relations in the world. In our simulations of the RO (relations only) task, the "Balint's" version of the DBO (dynamic binding only) architecture, which had representations of relations in its memory, performed no better than the "Cat" version of the architecture, which had relations neither in perception nor memory.

A third interesting finding was that carefully observing the temporal dynamics of individual units while LISA performs these reasoning tasks reveals the mechanisms whereby multi-place predicates and structure mapping perform the functions they do. Close examination of the RO and R&M architectures performing the RO task showed that the hierarchical structure of a multi-place predicate allows the architecture to overcome otherwise misleading semantic matches to the roles of a multi-place relation. Specifically, as visible in Figure 4, the semantic features of the critical predicates initially bias the system to match properties in perception to incorrect properties in memory. However, in the relational architectures (RO & R&M) it is possible to see this bias being overcome in real time in the time traces. Similarly, in the MO task, all architectures fail to infer the affordance of the first firing of the propositions in perception. However, by virtue of being able to learn structure mappings during the first firing of propositions, the MO & R&M architectures were able to make the correct inference upon the second firing of these propositions (as visible in Figure 5). Even in these architectures, however, you can see that they were initially biased toward making the wrong inference, but that the learned mappings quickly overcame this tendency.

Our conclusion is that the capacity to form multi-place predicates and the capacity to compute structure mappings, both instances of hierarchical integration of multiple role-bindings, are the minimal computational ingredients that separate symbolic neural computing architectures, such as the human brain, from the brains of otherwise intelligent organisms (such as bumblebees, cats, monkeys, chimpanzees) that nonetheless lack the capacity for symbolic thought and role-based relational reasoning (RRR). Armed with multi-place predicates and structure mapping, it appears that CSWG is also a necessary, but perhaps comparatively straightforward, additional ability required to benefit from the full advantages of symbolic thought.

**Relations Without Mapping and Mapping Without Relations**

In the 2x2 matrix of cognitive architectures and tasks explored in this paper, MO and RO may in some ways be the most interesting. Neither architecture gives rise to fully symbolic thought, but each appears to confer a useful approximation, and seems to map (speculatively) onto interesting cognitive abilities of some animals.

It is tempting to interpret bumblebees' behavior as reflecting an ability to respond to relations between objects, at least when those objects are sufficiently close together (Martin-Ordas, 2023). Blaisdell et al. (2006) presented evidence that rats represent causality as a relation, that is, as a structure that takes multiple arguments (the cause and the effect; see also Cheng, 1997). This result suggests that, in the domain of causal induction, rats (and presumably other animals) may be able to represent multi-place relations. As illustrated by our simulations of the MO (mapping only) task, Chimpanzees' tool use is consistent with the hypothesis that chimpanzees are capable of structure mapping, even if they are unable to represent multi-place relations (see Povinelli, 2003; Penn et al., 2008). Given the very limited nature of our review of the animal cognition literature, it seems safe to speculate that many other animals exhibit behaviors that demand explanation in terms of multi-place predicates, structure mapping, or both.

At the same time, bumblebees, rats, and chimpanzees have so far given us little reason to believe they are capable of fully symbolic thought. In our view, the jury is still out on cetaceans and corvids. If fully symbolic thought really does require both multi-place predicates and structure mapping, and possibly CWSG, then this reality may offer an answer to the question of why symbolic cognition took so long to evolve: Because it requires two, and perhaps three, mutations to occur in the same organism.

**The importance of CWSG**

Prior simulations (e.g., Doumas et al., 2008, 2022; Hummel, 2011; Hummel & Holyoak, 1997, 2003; Hummel et al., 2014) demonstrated that architectures capable of CWSG can also perform more complex symbolic tasks than even the R&M task explored here. To keep the mechanism of inference constant across our simulations, the simulations reported here were designed to work using an inference mechanism much simpler than CWSG, specifically, getting a semantic unit representing an affordance to fire in synchrony with the object to which it applied. In other words, the version of LISA used in the simulations was simpler than the version in Hummel and Holyoak (2003) because it was not permitted to use CWSG.

For the purposes of inferring basic affordances, the semantic inference metric used in the simulations reported here was adequate by design. It allows the cognitive architecture to infer a single property (e.g., "I can walk on this") about a single object or argument (e.g., a surface) simply by augmenting the predicate representing the object (here, surface) with a semantic unit representing the affordance. But for the purposes of symbolic thought more broadly, this form of inference clearly is not sufficient. People routinely make inferences about whole systems of objects and the relations among them, and these inferences typically take the form of systems of propositions over numerous arguments (rather than a single property or affordance of a single object).

**A Computational Phylogenesis of Biological Intelligence**

For biological intelligence, rapid generalization is the gold standard, so it is reasonable to speculate that evolution solved this problem, or at least the most basic version of it, hundreds of millions of years ago

**The Role of Invariants in Generalization** If one's goal is rapid generalization, then the first rule is to represent the meaningful properties of one's universe explicitly and independently of everything else, that is, as invariants. For example, if red is a reliable indicator of ripe fruit, and if you eat fruit for a living, then being able to detect when something is red, regardless of its other properties, is extremely adaptive. If "having large teeth and approaching me rapidly" is a reliable indicator that you are about to be attacked by a predator, then being able to detect when something has large teeth and is rapidly approaching you is also adaptive. If you are such an organism, then having neurons that respond to things like *red*, *big teeth*, and *coming-at-me* independently of everything else going on in the stimulus makes it possible to respond appropriately to new situations in which these properties are present. Such neurons can immediately help you to choose your next actions. But if you can only represent *red* and *coming-at-me* as components of various conjunctive (a.k.a., entangled) representations (e.g., *red-and-lower-left-visual-field-and-round-and…*), then choosing the appropriate action will depend, not on the presence of the crucial property (*red*), but on the crucial property occurring in a familiar conjunction with other properties. Such an organism may not live long enough to reproduce.

Representing the world in terms of invariants is necessary for the purposes of rapid generalization, but it is not sufficient to specify how the invariants present in any given stimulus go together to form meaningful objects (Hummel & Biederman, 1992). An animal looking at red berries while a predator is approaching them would detect the invariants *red*, *round*, *looming*, and *big teeth*, but would be unable to distinguish whether the red thing was also the round thing while the looming thing has big teeth from a situation in which the red thing is looming while the round thing has big teeth. For this reason, a representation based on invariants must have some basis for binding those invariants together dynamically, to distinguish *red+round* and *looming+big-teeth* from *red+looming* and *round+big-teeth*. As this limitation illustrates, the capacity to represent invariants, while essential for rapid generalization, is useless without the ability to bind those invariants dynamically into meaningful groups.

It is for this reason, we argue, that even fruit flies and bumblebees have evolved solutions to the problems of invariants and dynamic binding (see, e.g., Loukola et al., 2017; Tanaka, et al., 2009; Turner et al., 2007). Fruit flies need to be able to detect important invariants in their environment (e.g., the chemicals that suggest the presence of rotting fruit) and to know how those invariants go together to form meaningful stimuli (e.g., do the chemicals come from a common source?). We therefore take the ability to detect important invariants and the ability to bind those invariants dynamically into meaningful groups to be basic entry requirements for any biological intelligence.

**Generalization from Single-Place Predicates** Armed with the ability to represent important properties of their environment as invariants and dynamically bind them into meaningful groups, an animal is equipped to make universal inductive generalizations based on single-place predicates, as demonstrated in our Dynamic Binding Only (DBO) simulations. Even a cat can respond to important invariants in novel environments. This capacity strongly suggests that the cat's representations of the invariants in its environment are not entangled with anything else. And the fact that even young kittens have this ability suggests possessing it does not require extensive exposure to every possible combination of important properties in the world.

In the language of machine learning, the capacity to represent relevant properties (such as affordances) as invariants and bind them dynamically to new arguments "expands the convex hull" of training beyond the conjunctions of properties experienced in the training set. To a computational system without this capacity, it is necessary to learn many examples of an affordance as separate instances since every example is entangled with the context in which appears. The affordance, itself, is never explicitly represented, so the "convex hull" is restricted to the set of specific circumstances in which the affordance was previously experienced and interpolations between them.

By contrast, a system that can represent the affordance "can walk on" as an invariant (e.g., using neurons that respond specifically to that affordance, independent of everything else) and bind it to arbitrary arguments can walk on any walkable surface in any new environment, even statistically unlikely ones. Indeed, the greatest danger posed by such a representation is a propensity to overgeneralize: To a young enough child, every quadruped is a "doggie".

**Representing Relations** The capacity to bind single-place predicates to their arguments is sufficient to account for how people and other animals perceive basic affordances, but it is not sufficient to explain how we reason about the relations between things in the world.

A relational representation is a multi-place predicate of the form $r(x, y, \ldots z)$, that expresses a relation, $r$, between two (or more) arguments, $x, y \ldots z$. At first blush, it may appear that basic affordances take this form, where $r$ is the affordance, $x$ is the stimulus possessing the affordance, and $y$ is the observer, as in the affordance *can-walk-on* (that-surface, me). However, in the case of basic affordances, the second argument is always either *me* or (more likely) some part of me such as *my right hand*, *my eye*, *my left foot*, etc. For this reason, basic affordances can be reduced to single-place predicates, such as *can-walk-on* ($x$), or *can-grasp* ($x$), where the argument, $x$, is bound dynamically to the object (e.g., *this-surface* or *this-object*), but the second argument is absorbed into the predicate itself (in the case of *can-walk-on*, the absorbed argument is *me*, whereas in the case of *can-grasp*, it might be *my-right-hand*).

We argue that this is why even animals without the capacity for relational representations (including fruit flies, cats, and monkeys, all the way up to chimps and bonobos; see Penn, et al., 2008) can make their way in the world but generally cannot reason about the relations between objects exterior to themselves.

However, to manipulate the world as people do, it is necessary to understand the relations between various objects in the world and be able to reason about those relations as entities in their own right. The capacity to represent and reason about relations as explicit, independent entities is evident in our ability to reason using schemas and analogies. It is trivial for most adults to understand the analogy between the structure of the solar system and the Rutherford model of the atom, but our ability to do so rests on our understanding that the *orbit* ($x, y$) relation is the same thing in the context of *orbit* (planets, sun) as in *orbit* (electrons, nucleus).

Recall that armed with a capacity for recursion, a representational system with a finite number of basic elements can represent an unbounded number of specific structures. The effect of relational representations on our capacity for generalization scales accordingly. If being able to dynamically bind invariants to one another makes it possible to generalize to new situations containing those invariants (e.g., by making it possible to generalize from, say, a single exposure to a tasty red berry to the inference that all red things are tasty), being able to represent relational structures explodes the convex hull, in part by making it possible to represent properties of the universe that cannot be expressed at all with single-place predicates.

**Mapping-based Generalization** The ability to represent relations gives a representational system unbounded expressive power but being able to map one relational structure onto another makes it possible to use one structure as a representation of another. Analogical reasoning is an instance of using the source analog as a representation of the target. For example, Gick & Holyoak's (1980, 1983) subjects used a story of a general sending troops to capture a fortress in the center of a city from multiple directions (the source analog) to infer that a doctor could use radiation converging from multiple directions to destroy a tumor (the target analog). In this analogy, the subjects are representing the radiation problem in terms of the story about the general, where the general represents the doctor, the fortress represents the tumor, the town represents the healthy tissue, and the troops represent the radiation. Using the general story to reason about the radiation problem in this way is not unlike using an equation to solve a problem in physics. In both cases, the relevant parts of the target are mapped to the relevant parts of the source, and the structure of the source dictates the solution to the target. To date, these capacities have only been observed in humans (Penn et al., 2008).

Reasoning about one system in terms of another, better understood, system effectively brings everything that is known about the better-known system to bear on the problem of understanding the less well-known system. This process is akin to expanding the smaller convex hull (the less well-known system) by merging it with the convex hull of the better-known system to form a larger effective convex hull. The power of this capacity to expand the scope of our reasoning is evidenced by the ubiquity of analogical thinking in human mental life (Holyoak & Thagard, 1995). Whenever we make an analogy between two structures, we induce a more general schema specifying what the original systems have in common and deemphasizing the details on which they differ (see Gentner, 1983; Gick & Holyoak, 1980, 1983; Holyoak & Thagard, 1995). When we discover an analogy between two seemingly very different things, it is as though we have discovered a greater, more general truth about the universe.

**Language**   In this paper, we have focused exclusively on inference and generalization, up to and including relational reasoning, particularly in the context of vision. But it is worth noting that both multi-place relations and mapping play essential roles in language. Accordingly, it is tempting to speculate that the principles we have discussed here may be relevant to the origins of language (see also Deacon, 1997).

All languages express multi-argument relations, permitting recursion, which has been observed in all human languages, with one possible exception (Everett, 1986). As a representational system, any language must be mappable onto representations of the ideas expressed by the language. This is not to say that these mappings are always explicit, as they are in the case of analogical reasoning, but they are at least capable of being made explicit. Asked about the correspondences between, say, an image and a verbal description of that image, most people can point out which parts of the text refer to which parts of the image. A sentence is a representation of whatever it is about, so there is necessarily a mapping between parts of the sentence and aspects of the thing being represented, even if some of these mappings are not obvious. It is therefore at least possible that the same mutations that gave rise to our capacity for multi-place relations and structure mapping also played a central role in the evolution of language.

**The Relationship Between Biological and Artificial Intelligence**

The apparent successes of transformer architectures would seem to call every one of our claims into question: *Is dynamic binding necessary? Large Language Models (LLMs) don't do dynamic binding, and yet they converse like a person. Are multi-place predicates necessary? LLMs don't form multi-place predicates, and yet they know more than anyone I've ever met. Is structure mapping necessary? LLMs don't do structure mapping, and yet they succeed on many of the tasks you cite as requiring symbolic thought.*

Indeed, transformer models seem to obviate all these things — dynamic binding, multi-place predicates, and even structure mapping and CWSG — via their attention heads and extensive training. Webb et al. (2023) showed that some LLMs appear to solve at least some analogies, which would suggest some capacity to at least approximate symbolic thought. These successes cast doubt on the need for explicit structure mapping in analogical reasoning. However, it worth pointing out that it is not well understood exactly how these models solve analogies. It is at least plausible that the immense quantity of text on which LLMs are trained makes it possible for them to solve such problems based only on the statistics of their training corpora. It is also possible that the dynamic weighting mechanisms in a transformer are similar to, or a substitute for, dynamic binding in biological mechanisms[8].

However, transformers often struggle to correctly bind representational elements into statistically unusual relations, in part because they are dependent on conjunctions of properties in the examples on which they have been trained. As a result, relational roles are not represented independently of their fillers, rendering the system dependent on familiar role-filler combinations. The effect of entanglement on neural networks can be visually observed by asking a transformer model to create an image of an arrangement of objects with statistically uncommon role-filler bindings. Such a request can cause the network to make nonsensical images, where objects that should be separate become physically connected, or the transformer may just depict the objects in a more statistically likely arrangement instead.

Take for example the prompt, "Can you create a photorealistic image in a 1:1 aspect ratio of a new daytime viewpoint of a tent erected on the wing of a plane which is parked on the wing of seagull?" Three state-of-the-art networks at the time of this writing (ChatGPT o4-mini-high, OpenAI, 2025; ChatGPT o3, OpenAI, 2025; Gemini 2.5 Flash, Google, 2025) showed that they struggled to represent seagulls, planes, and tents independently of each other and in the correct relations. As shown in Figure 8, the networks hybridized the airplanes and seagulls, put objects in the wrong relationships to each other or on the wrong parts of objects, removed parts of objects or added disembodied parts, and they struggled to individuate separate tokens of the same kind of object (that is, they confused the wings of the airplane with the wings of a seagull). People, by contrast, do not struggle to imagine what such a scene would look like, although it is interesting to wonder whether the DBO and MO architectures would.

---

[8] Transformers dynamically change connection weights between units, which has been likened to attention, but these weight changes are also a function of the training set. For this reason, transformers do not achieve the kind of fully flexible dynamic binding discussed here.

## ChatGPT o4-mini-high

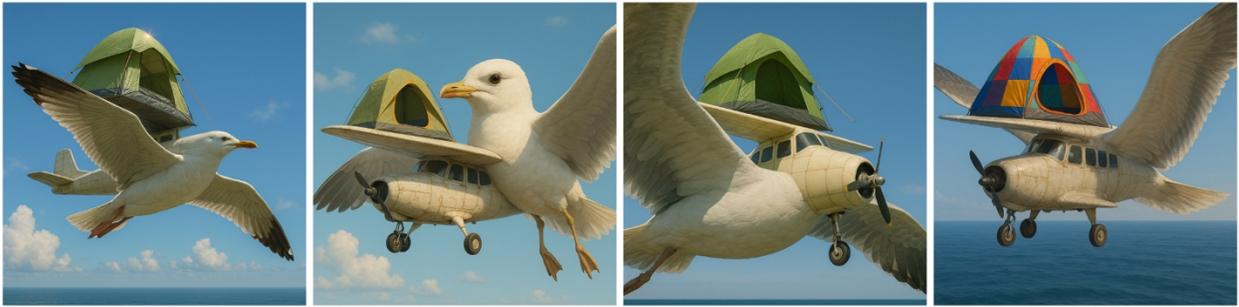

## ChatGPT o3

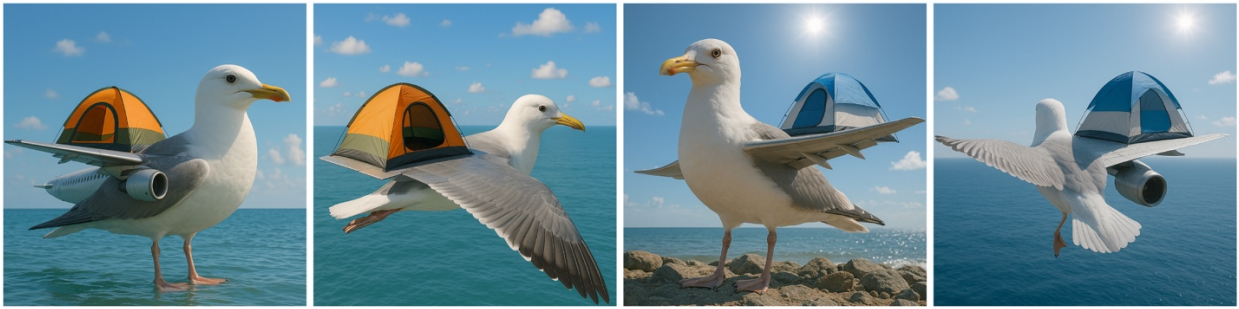

## Google Gemini 2.5

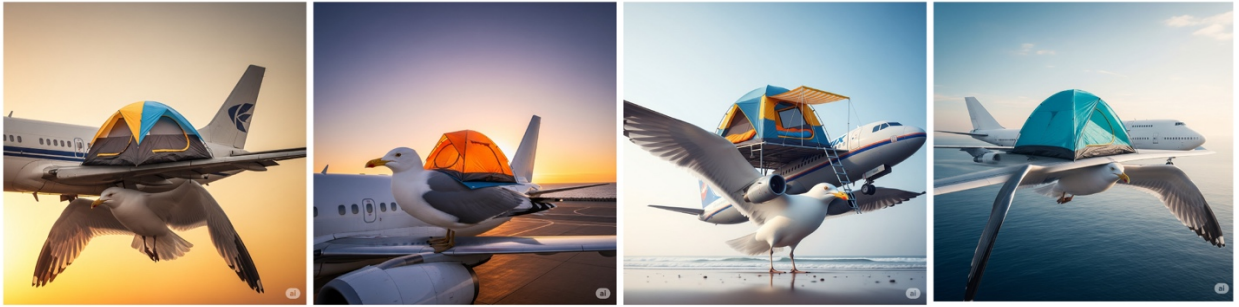

**Figure 8**

Images produced by three networks in response to the prompt, "Can you create a photorealistic image in a 1:1 aspect ratio of a new daytime viewpoint of a tent erected on the wing of a plane which is parked on the wing of seagull?" First row: ChatGPT o4-mini-high (OpenAI, 2025), OpenAI's model that specializes in visual reasoning, tends toward hybridizing the plane and the bird. Second row: ChatGPT o3 (OpenAI, 2025), OpenAI's advanced reasoning model, also tends to hybridize the plane and the bird. Third row: Google Gemini 2.5 Flash (Google, 2025), displays some tendency toward hybridization, but also reverses relations or adds duplicates of parts when generating images of unusual spatial relations. The first image in each row depicts the network's best response to the prompt.

The complete set of generated images (the first 12 images generated by each network in response to the prompt) is available at https://osf.io/p5zjw.

Finding errors like the examples above will always be a moving target, because as soon as the network is trained with an example that matches the prompt, the conjunction of properties will reside within the convex hull of the network's training set. While the specific example will no longer give rise to an error, the class of errors will persist.

The result of these systems' reliance on entangled representations is that they require an enormous amount of training to perform any task. Being limited, for example, to representing "this shape in this location in the visual field, at this size, in this color, in this viewpoint, in this relational role" means that, to recognize a given object in any location, size, color, viewpoint, or relationship to other objects, the networks must effectively be trained on every object in every possible location, size, color, viewpoint, and relationship (see Montero et al., 2020; 2022; Schott et al., 2021) in order to try to expand the convex hull. LLMs as models of human cognition must be trained on a body of text that it would take a human thousands of years to read, even assuming said human did nothing but read the entire time.

A human, by contrast, learns to converse intelligently within about three years. Children learn words in at most two or three examples (Smith & Yu, 2008). People can also recognize a novel object at any location, size, color, viewpoint, and relationship to other any other object after just a single viewing (barring "accidental" views that make it impossible to recover the object's shape; Biederman, 1987). People extrapolate as a matter of course. Clearly, biological intelligence comes equipped with capacities that do not depend on extensive training and are not limited to the convex hull of the training set. This capacity is true of animals whether or not they have the capacity of for symbolic thought.

One of the advantages of the biological approach appears to be resource efficiency. If one of the goals for artificial intelligence is to create a system that behaves like humans and runs on the power of a single light bulb rather than producing the carbon footprint of New York City operating for a month, it may be useful to look for inspiration from biological intelligence.

Our goal in this paper was to explore how biological intelligence achieves symbolic cognition. LISA, while not an image computable model, is an explainable, biologically-inspired neural model of human reasoning. In the simulations reported here, the only prior examples on which LISA was "trained" were the Memory analogs. In other words, the DBO, MO, RO, and R&M architectures explored here each made the inferences they did based on only a single training example. Principles from the LISA architecture could be useful for developing artificial neural networks that, like people and other animals, are less resource intensive and require less training.

**Conclusion**

Our simulations results suggest that the difference between most biological intelligence and human intelligence is that we, possibly along a few other notable examples, evolved the ability to integrate multiple dynamic role bindings into multi-place relations along with the ability to map systems of such relations onto one another. Together, these abilities make it possible to (a) represent and reason about the relations among multiple entities (either objects or relations) in the universe, (b) form recursive structures that qualitatively expand the capacity of our mental representations, and (c) map representations onto one another, making it possible for us to use the symbolic representations we can generate.